%% file: main.tex
\documentclass[10pt,twocolumn,letterpaper]{article}

\usepackage{iccv}
\usepackage{times}
\usepackage{epsfig}
\usepackage{graphicx}
\usepackage{grffile}
\usepackage{amsmath}
\usepackage{amssymb}
\usepackage{booktabs}

\usepackage{enumitem}

\usepackage[pagebackref=true,breaklinks=true,letterpaper=true,colorlinks,bookmarks=false]{hyperref}

\iccvfinalcopy 

\newcommand\comment[1]{}
\newcommand\todo[1]{}
\newcommand\degr[0]{^{\circ}}

\newcommand{\boldstartspace}[1]{\vspace{0.1in}\noindent\textbf{#1}}

\def\iccvPaperID{5073} 

\ificcvfinal\fi
\begin{document}

\title{Deep Parametric Indoor Lighting Estimation}

\author{Marc-Andr\'e Gardner\textsuperscript{*}, \hspace{3em}Yannick Hold-Geoffroy\textsuperscript{$\dagger$}, \hspace{3em}Kalyan Sunkavalli\textsuperscript{$\dagger$},\\
Christian Gagn\'e\textsuperscript{*}, \hspace{3em}Jean-Fran\c cois Lalonde\textsuperscript{*}\\
\vspace{0.4em}
\textsuperscript{*}Universit\'e Laval, \textsuperscript{$\dagger$}Adobe Research\\
{\tt\small marc-andre.gardner.1@ulaval.ca \hspace{3em}\{holdgeof,sunkaval\}@adobe.com}\\
{\tt\small \{christian.gagne,jflalonde\}@gel.ulaval.ca}\\
{\tt\small \url{https://lvsn.github.io/deepparametric/}}%
}

\maketitle

\begin{abstract}
   We present a method to estimate lighting from a single image of an indoor scene. Previous work has used an environment map representation that does not account for the localized nature of indoor lighting. Instead, we represent lighting as a set of discrete 3D lights with geometric and photometric parameters. We train a deep neural network to regress these parameters from a single image, on a dataset of environment maps annotated with depth. We propose a differentiable layer to convert these parameters to an environment map to compute our loss; this bypasses the challenge of establishing correspondences between estimated and ground truth lights. We demonstrate, via quantitative and qualitative evaluations, that our representation and training scheme lead to more accurate results compared to previous work, while allowing for more realistic 3D object compositing with spatially-varying lighting.  
\end{abstract}

\input{intro}
\input{related}
\input{method}
\input{results}
\input{discussion}

\section*{Acknowledgments}

We acknowledge the financial support of NSERC for the main author PhD scholarship. This work was supported by the REPARTI Strategic Network, the NSERC Discovery Grant RGPIN-2014-05314, MITACS, Prompt-Qu\'ebec and E Machine Learning. We gratefully acknowledge the support of Nvidia with the donation of the GPUs used for this work, as well as Adobe with generous gift funding. 

{\small
\bibliographystyle{ieee_fullname}
\bibliography{bibliography}
}

\end{document}

%% file: intro.tex
\section{Introduction}
\label{sec:introduction}

Recovering the lighting in a scene from a single image is a highly ill-posed problem. Since images are formed by conflating lighting with surface reflectance, scene geometry and the camera response function, inverting the image formation process to recover any of these components is severely under-constrained. This is especially true when the image has low dynamic range (LDR) and limited field of view, such as one captured by standard consumer cameras. 

\begin{figure}[!t]
    \centering
    \footnotesize
    \setlength{\tabcolsep}{1pt}
    \includegraphics[width=\linewidth]{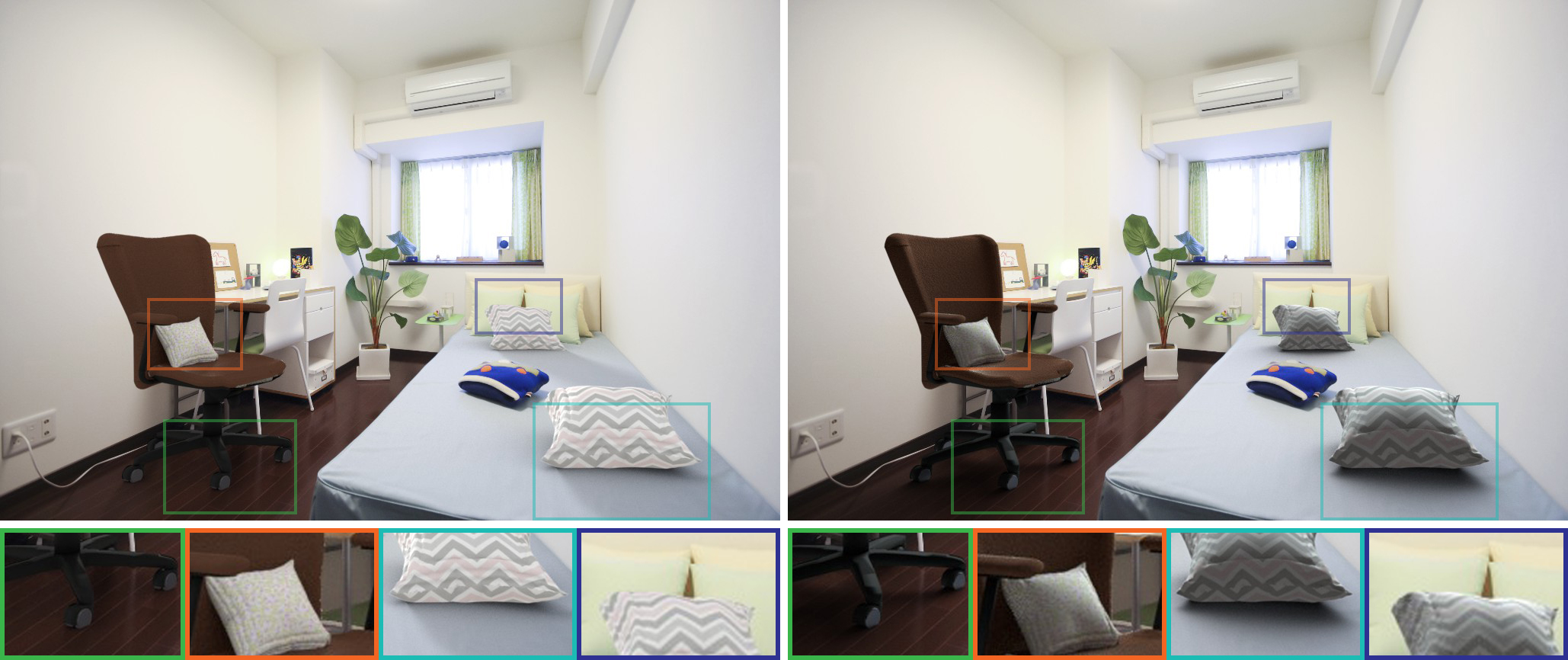}
    \begin{tabular}{@{}p{0.5\linewidth}@{}p{0.5\linewidth}@{}}
        \hspace{1cm} (a) Gardner et al.~\cite{gardner-sigasia-17} &
        \hspace{1.3cm} (b) Ours \\
    \end{tabular}
    \caption{Our method takes an indoor image as input and estimates a parametric lighting representation, enabling tasks such as virtual object insertion. Unlike previous methods that predict global lighting~\cite{gardner-sigasia-17} (a), our estimates vary based on spatial location (b), generating different shadow directions (green, teal) and shading (orange, blue) that realistically adapt to the location in the image.}
    \label{fig:teaser}
\end{figure}

In his pioneering work in image-based lighting, Debevec~\cite{debevec-sig-98} proposed to directly capture the lighting conditions at a location in the image by inserting a light probe at that location. The resulting HDR environment map represents the illumination incident from every direction \emph{at that point in the scene}, and can be used to realistically relight virtual objects at that location. Recently, fully automatic methods leveraging deep learning were proposed to estimate an environment map from a single indoor image~\cite{gardner-sigasia-17}. However, the environment map representation assumes that lighting is distant --- which is why it can be represented as a function of the incident direction. This assumption is often violated for indoor scenes that have localized light sources leading to spatially-varying lighting in the scene. Consequently, using a single environment map for the entire scene leads to inconsistent results for applications such as 3D object compositing (see Figure~\ref{fig:teaser}(a)).

Alternatively, \emph{parametric} lighting models represent illumination using a discrete set of light sources with various geometric and photometric properties. This is a \emph{global} representation that can be used to reconstruct lighting at \emph{any} 3D location in the scene (assuming known scene geometry). Techniques for learning parametric lighting have been proposed for the simpler case of \emph{outdoor} illumination~\cite{lalonde-ijcv-12,holdgeoffroy-cvpr-17}. Indoor lighting, on the other hand, is much more complex, with a varying number of light sources, all with potentially different properties, and located in close proximity to the scene. As a result, techniques for recovering parametric lighting in indoor scenes rely on extensive user input~\cite{karsch-siga-11} or hand-crafted heuristics that can often fail~\cite{karsch-tog-14,barron2013rgbd}.

In this paper, we propose a learning-based method for estimating parametric lighting from a single indoor image. In particular, our method produces a set of lighting parameters describing area lights \emph{distributed in 3D}, from a single 2D LDR image. Specifically, our method predicts the light source positions (in 3D), areas, intensities and colors. These parameters can be directly plugged into a rendering engine to relight an object at any location in the scene, leading to photorealistic results as in Figure~\ref{fig:teaser}(b). Moreover, because these parameters are directly related to the physical properties of lighting, they provide an intutitive space for artists to manipulate and design the lighting they want for the scene.

Our network is trained on a large dataset of indoor HDR environment maps~\cite{gardner-sigasia-17} that we have manually labelled with pixel-wise depth information and annotated with our parametric lights. We encode an input RGB image into a latent feature vector from which we decode the parameters for a fixed number of light sources. Compared to Gardner et al.~\cite{gardner-sigasia-17} who regress a full environment map, this is a significantly smaller set of parameters, leading to better performance while training on a much smaller dataset. 

One approach to train our network would be to penalize the differences between the predicted lights and the annotated lights. However, this requires establishing a correspondence between these two sets of lights. This is fragile, especially earlier in training when the estimated light sources are arbitrary, and is unstable to train. Instead, we utilize a differentiable (non-learnable) layer that converts the predicted light parameters to an environment map. This allows us to compute the loss directly with respect to the environment maps without requiring any correspondences between light sources. Once trained in this manner, the network predicts good positional information, which we use to establish correspondences and further fine-tune the parameters. We demonstrate that this yields illumination estimates that are superior to previous lighting estimation methods. 


\boldstartspace{Contributions.} In summary, our main contributions are:
\begin{enumerate}[nosep,leftmargin=*]
	\item A deep neural network to estimate parametric 3D lighting from a single indoor image; 
	\item A dataset of depth annotations over panoramas that can be used to learn 3D scene lighting;
    \item A robust training procedure based on a differentiable parametric loss with respect to environment maps.
\end{enumerate}

%% file: related.tex
\section{Related Work}
\label{sec:related}

Illumination estimation is a classic vision/graphics problem and is a critical component of work on scene reconstruction and rendering. A large body of work on illumination estimation focuses on reconstructing lighting from images of individual objects. Some of these methods assume that the shape of the object is either known~\cite{marschner1997inverse,lombardi2016reflectance} or can be reconstructed using shape priors~\cite{blanz1993dmm} or heuristics~\cite{moreno-cgf-13}. Barron and Malik~\cite{barron-pami-15} recover geometry, reflectance and illumination from a single image of an arbitrary object by enforcing hand-crafted priors on each component. Recently, deep learning-based methods have been proposed to recover illumination and material properties (along with, in some cases, geometry) from a single RGB image of an object~\cite{georgoulis-iccv-17,liu2017material,meka-cvpr-18,li-sig-2018}. These methods do not easily scale to large-scale indoor scenes where the illumination, geometry, and reflectance properties are significantly more complex. 
 
Methods for estimating lighting in large-scale indoor scenes often assume known geometry. Barron and Malik~\cite{barron2013rgbd} and Maier et al.~\cite{maier2017intrinsic3d} assume an RGBD input image and recover spatially-varying Spherical Harmonics illumination. Zhang et al.~\cite{zhang-siga-16} reconstruct parametric 3D lighting but require a full multi-view 3D reconstruction of the scene. Karsch et al.~\cite{karsch-siga-11} recover parametric 3D lighting from a single image, but do so by requiring substantial user input to reconstruct coarse geometry and initialize the lighting.  

Lalonde et al.~\cite{lalonde-ijcv-12} uses hand-crafted priors based on scene cues such as shadows to recover lighting from a single outdoor image. Hold-Geoffroy et al.~\cite{holdgeoffroy-cvpr-17} propose a deep neural network that does the same. Both of these methods utilize low-dimensional analytical outdoor illumination models; indoor illumination is significantly more complex. 

Similar to us, Karsch et al.~\cite{karsch-tog-14} estimate parametric 3D lighting from a single indoor image. They estimate scene geometry and reflectance, detect light source positions, and then estimate light source intensities using a rendering-based optimization. Each subtask in their pipeline is a challenging inverse problem that is solved using heuristics; errors in each component can propagate forward leading to inaccurate lighting estimates. In contrast, Gardner et al.~\cite{gardner-sigasia-17} propose an end-to-end deep network to regress lighting, represented as an environment map, from a single image. They train this network on a large-scale dataset of LDR environment maps~\cite{xiao-cvpr-12} and fine-tune it on an HDR environment map dataset. Legendre \etal extend this work to mobile applications and obtain better results by using a collection of videos as training data~\cite{LeGendre_2019_CVPR}. Song \etal~ \cite{Song_2019_CVPR} use Matterport3D~\cite{chang17matterport3D} dataset and a novel warping procedure in order to support multiple insertion points. We improve on their work by training an end-to-end neural network to predict discrete parametric 3D lights with 3D position, area, color and intensity. We show that this lighting model comes with many advantages. First, it is a compact representation that is easier to learn; we show that training only on the HDR environment map dataset leads to more accurate lighting predictions than their work. Second, a 3D representation naturally handles spatially-varying lighting at different scene points. Finally, we can give artists access to individual light source parameters, allowing them to intuitively edit scene lighting. 


%% file: method.tex
\section{Method}
\label{sec:method}

\begin{figure*}
    \centering
    \footnotesize
    \setlength{\tabcolsep}{2pt}
    \begin{tabular}{ccccc}

    \includegraphics[height=0.65in]{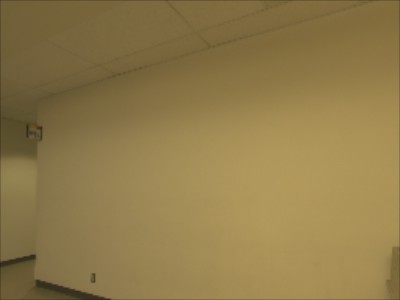} & 
    \includegraphics[height=0.65in]{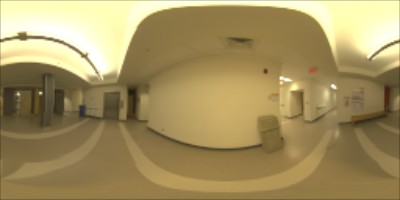} & 
    \includegraphics[height=0.65in]{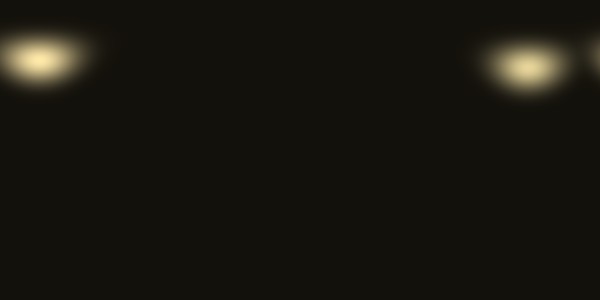} & 
    \includegraphics[height=0.65in]{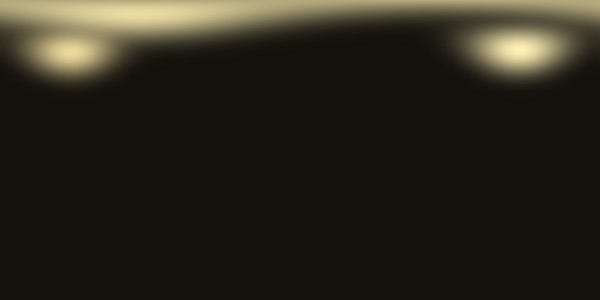} & \includegraphics[height=0.65in]{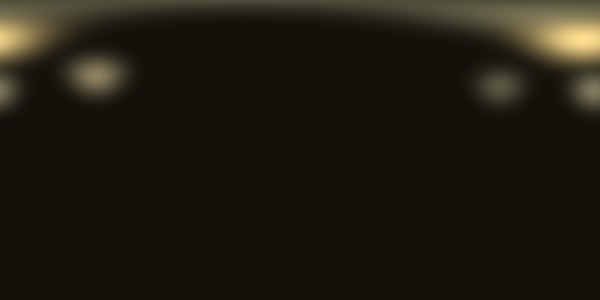} \\
    
    \includegraphics[height=0.65in]{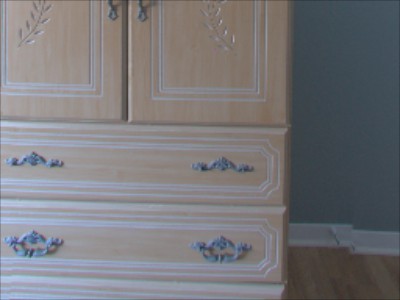} &
    \includegraphics[height=0.65in]{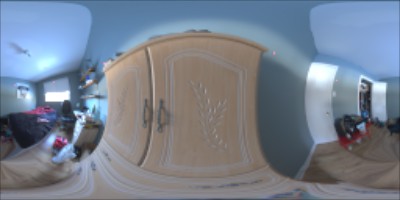} &
    \includegraphics[height=0.65in]{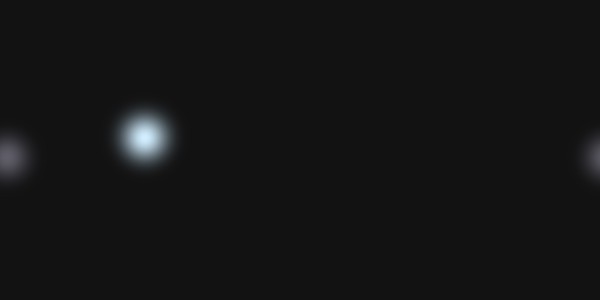} & 
    \includegraphics[height=0.65in]{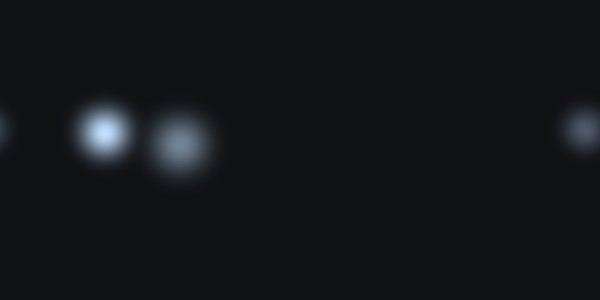} &
    \includegraphics[height=0.65in]{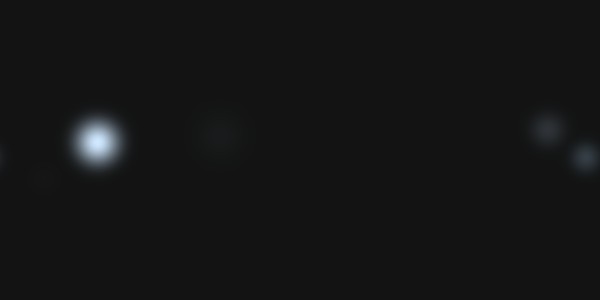} \\ 

    (a) Input & (b) GT Environment map & (c) 2 lights & (d) 3 lights & (e) 5 lights \\
    
    \end{tabular}
    \caption{Lighting predictions for networks trained with different numbers of lights $N$. From left to right: input image, ground truth lighting, and predictions for networks trained with $N=\{2,3,5\}$ respectively. When trained to output more lights, the network is able to effectively turn off some lights by decreasing their intensity in order to fit a scene requiring a lesser number of light sources.}
    \label{f:nbrlights}
    \vspace{-0.5em}
\end{figure*}

We aim to predict lighting conditions from a single, low dynamic range image. We frame this as the following learning problem: given input image $\mathcal{I}$, predict a set of parameters $\mathcal{P}$ which accurately represents the illumination conditions in $\mathcal{I}$. Let $\mathcal{P}$ be a set of $N$ lights and an ambient term:
\begin{equation}
    \mathcal{P} = \{\mathbf{p}_1, \mathbf{p}_2, \ldots, \mathbf{p}_N, \mathbf{a}\},
\end{equation}
where $\mathbf{a}\in \mathbb{R}^3$ is the ambient term in RGB. Each light $\mathbf{p}_i$ is represented by four parameters:
\begin{equation}
    \mathbf{p}_i = \{\mathbf{l}_i, d_i, s_i, \mathbf{c}_i\},
\end{equation}
where $\mathbf{l}_i \in \mathbb{R}^3$ is a unit vector specifying the direction of the light in XYZ coordinates (we found that encoding direction as a 3-vector was more stable to train compared to spherical coordinates that have problems with wrap-around), $d_i$ is a scalar encoding the distance in meters, $s_i$ the angular size of the light in steradians, and $\mathbf{c}_i \in \mathbb{R}^3$ the light source color in RGB. Here, $\mathbf{l}_i$, $d_i$ and $s_i$ are defined with respect to the camera. While $N$ may vary from one image to another, in practice we set $N$ to a fixed value (e.g. $N=3$) since it is always possible to effectively ``remove'' a light by setting $\|\mathbf{c}_i\| = 0$. We demonstrate this in our results in fig.~\ref{f:nbrlights}.

\begin{figure}
    \centering
    \includegraphics[width=0.495\linewidth]{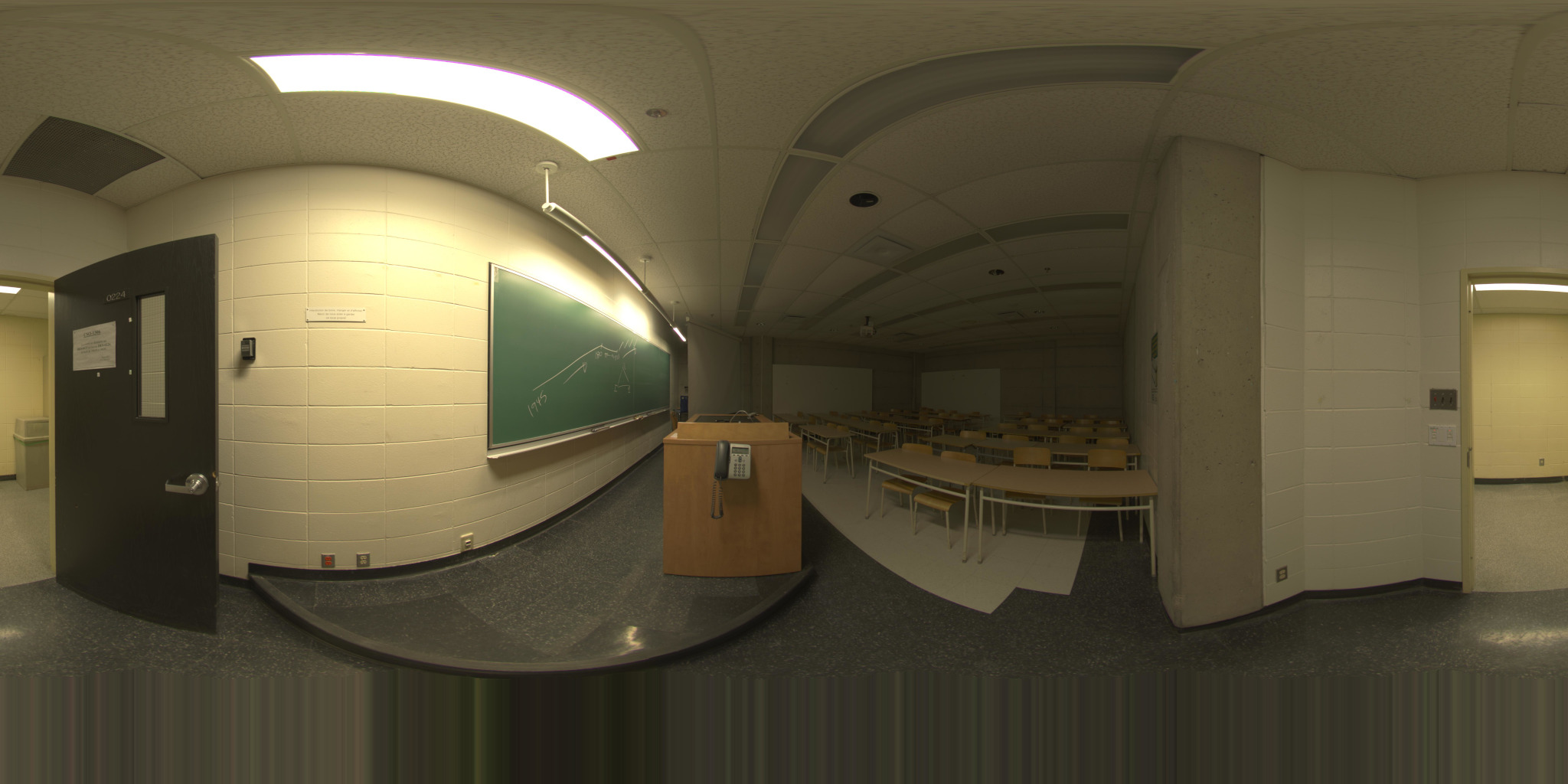} 
     \includegraphics[width=0.495\linewidth]{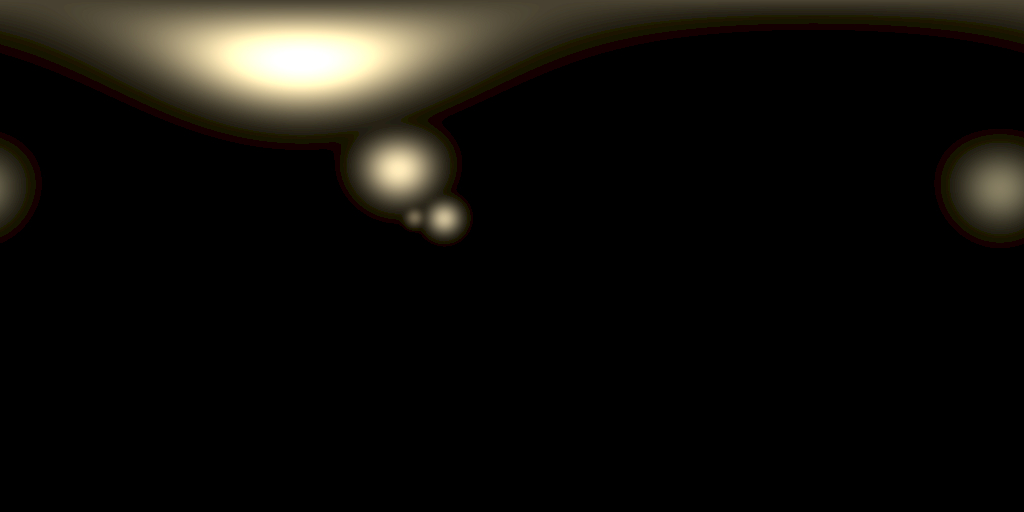}
     
     \includegraphics[width=0.495\linewidth]{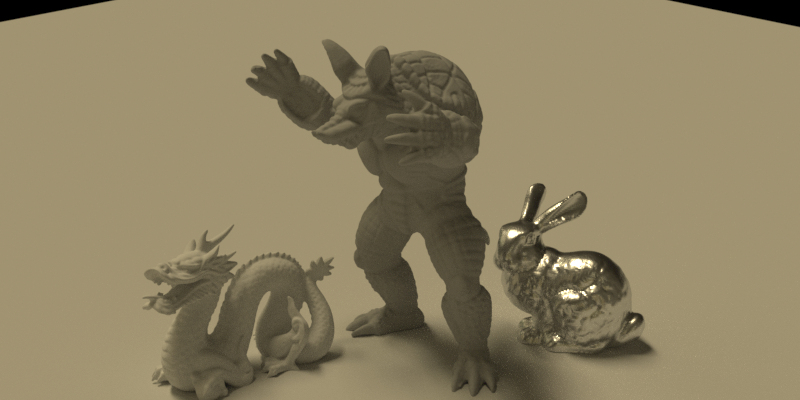}
     \includegraphics[width=0.495\linewidth]{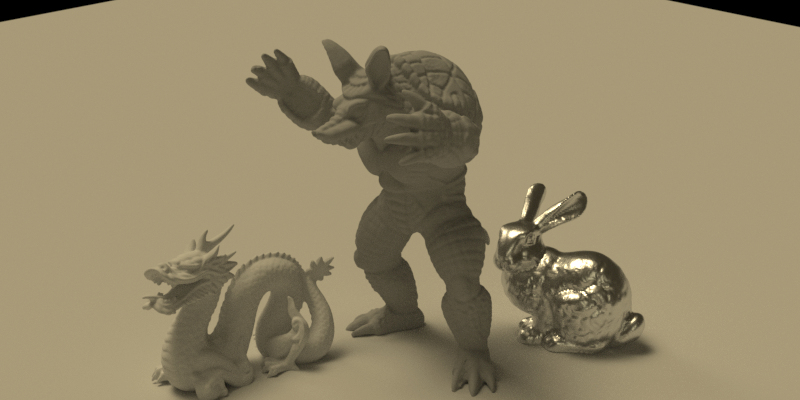}
     
    \includegraphics[width=0.495\linewidth]{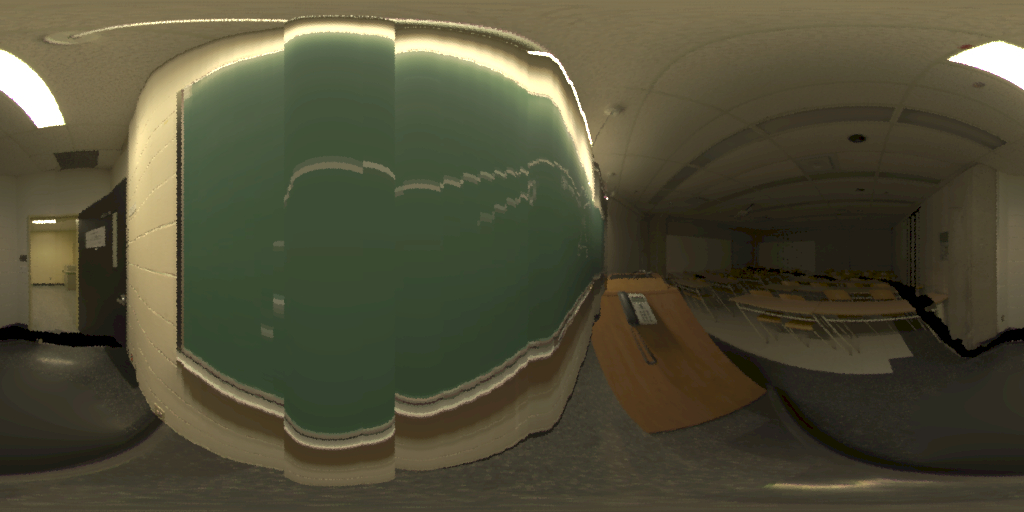} 
     \includegraphics[width=0.495\linewidth]{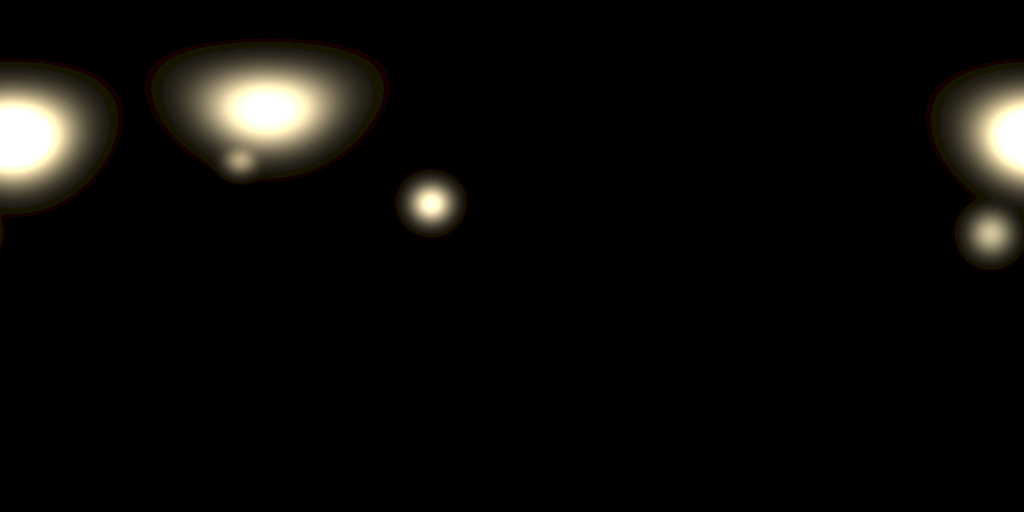}
     \includegraphics[width=0.495\linewidth]{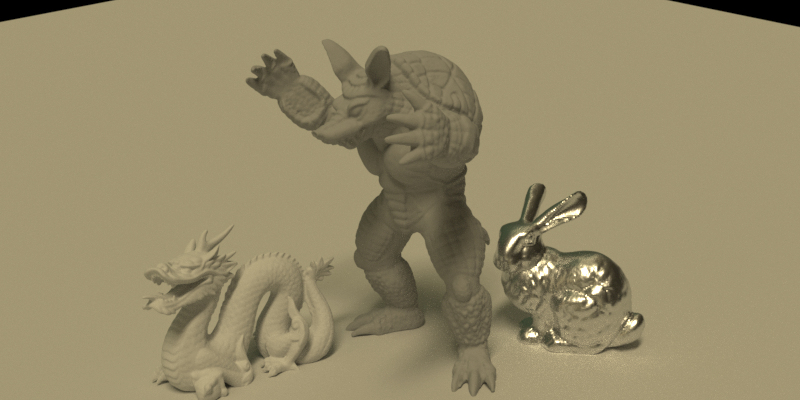}
     \includegraphics[width=0.495\linewidth]{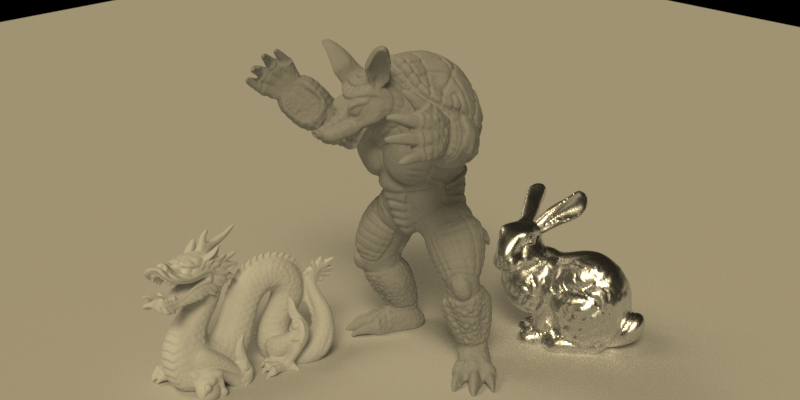}
    \caption{Comparison between an environment map and our lighting representation for relighting. Top row: source HDR panorama, and spherical gaussian representation of the parameters. Second row: renders generated by both representations. Third and fourth row: lighting and renders when moving in the scene, using EnvyDepth~\cite{banterle-cgf-13} as ground truth. Note how our parametric lights stay true to the ground truth for a different location in the scene.}
    \label{fig:parametric}
\end{figure}

    %
    %

\subsection{A Parametric Indoor Lighting Dataset}
\label{sec:dataset}

To train a deep lighting estimation network, we would ideally need a large dataset of images with labelled ground truth 3D light sources. Unfortunately, no such dataset exists. While the Matterport 3D dataset~\cite{chang17matterport3D} has a diverse set of indoor scenes with HDR panoramas, the tops of these panoramas (up to $30\degr$ from the zenith) are missing; given that a significant portion of indoor illumination comes from the ceilings, this is an issue for lighting estimation methods. Another option would be to use synthetic indoor scene datasets such as SUNCG~\cite{song2016ssc}; however, SUNCG has unrealistic material maps that do not match real world appearance. Instead, we rely entirely on real data for training. For this, we use the Laval Indoor HDR Dataset~\cite{gardner-sigasia-17}\footnote{Available at \url{http://indoor.hdrdb.com}.}, which contains 2,100 HDR panoramas taken in a variety of indoor environments. We manually annotated each panorama in this dataset using EnvyDepth~\cite{banterle-cgf-13} to obtain per-pixel depth estimates. As with \cite{gardner-sigasia-17}, we extract eight limited field-of-view crops for each panorama to form our dataset. 

We retrieve the ground truth light sources $\mathcal{P}$ in these panoramas with a simple algorithm. Here, there is no need to train light source detectors as in \cite{gardner-sigasia-17,karsch-tog-14} since the dataset is HDR. We first extract the peak value of the panorama. Then, simple region detection is employed to initialize seeds, which are then grown until the intensity goes under a third of the peak. Other lights are detected by repeating the process after masking the detected sources, until an energy threshold is met. The set of light parameters $\mathcal{P}$ is obtained by computing the corresponding $\mathbf{p}_i$ on each light source independently: $\mathbf{l}_i$ is the vector from the camera position pointing towards the center of mass of the light pixels, $d_i$ is the mean depth obtained from EnvyDepth, $s_i$ is the average angular size of the major and minor axes of an ellipse fitted on the light and $\mathbf{c}_i$ is the mean RGB color of the light pixels. Fig.~\ref{fig:parametric} shows the result of applying this algorithm on real HDR panoramas from the Laval Indoor HDR Dataset, and illustrates that doing so results in relighting results that are close to those obtained with the input environment map. 


We then process the dataset to retrieve $\mathcal{P}$ for all images. At this step, $N$ varies depending on the content of each environment map, ensuring that $\mathcal{P}$ models every significant light source (a light is deemed as significant if it is providing at least 10\% of the energy of the strongest source). Once extracted, light intensities are fine-tuned using a rendering-based optimization process. We render a simple object using the environment map masked everywhere except for the detected lights. We then render the same object using each of our extracted parametric lights. We optimize the parametric light colors and intensities such that they minimize the error with respect to the ground truth environment map.

\subsection{Network architecture}

\begin{figure}[t]
    \centering
    \includegraphics[width=.85\linewidth]{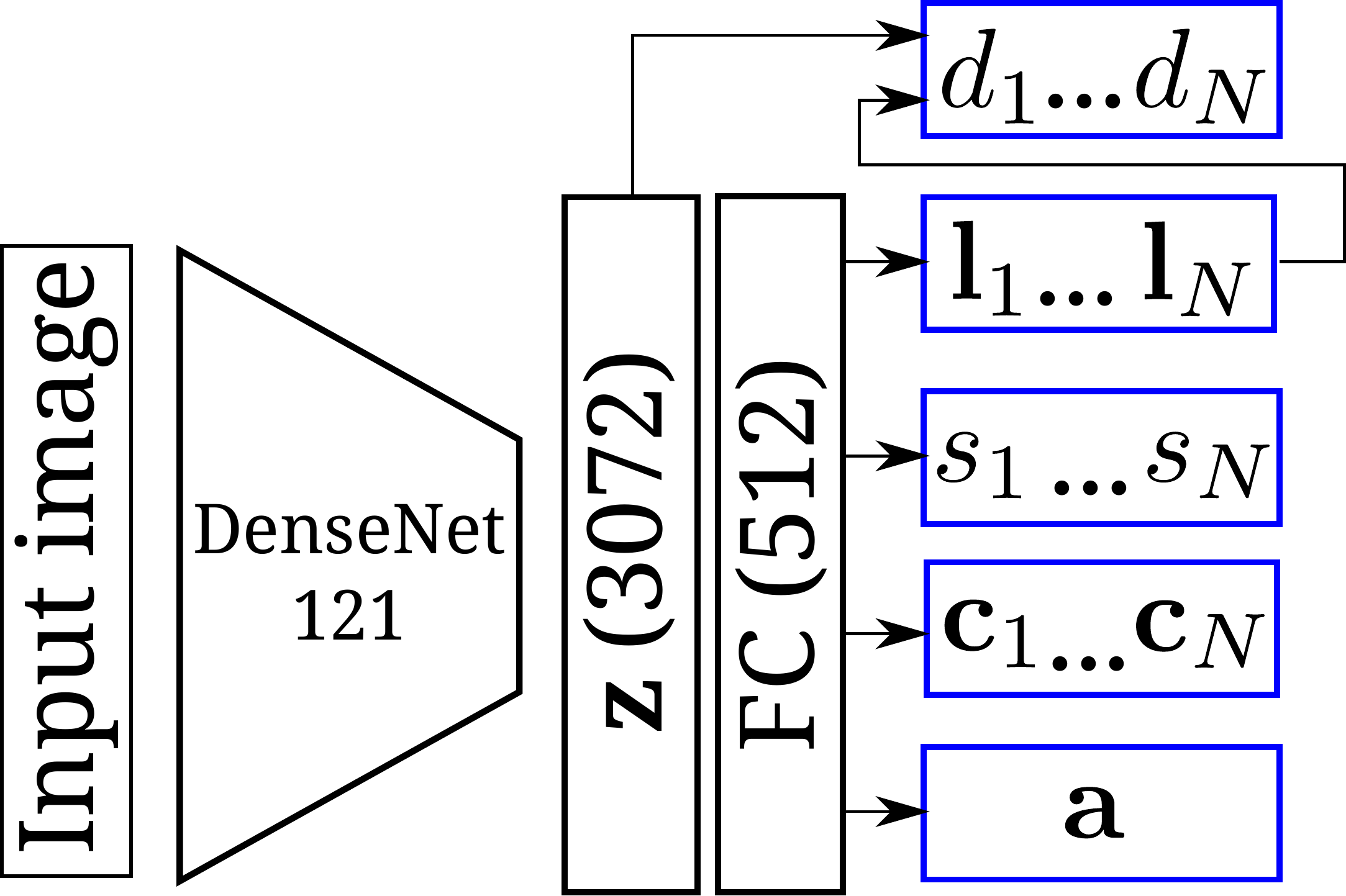}
    \caption{Proposed network architecture. A blue box indicates an output layer. Output sizes vary depending on $N$, except for $\mathbf{a}$.}
    \label{f:netarch}
\end{figure}

We learn the mapping from input image $\mathcal{I}$ to a set of estimated lighting parameters $\hat{\mathcal{P}}$ using a deep neural network, whose architecture is shown in fig.~\ref{f:netarch}. Specifically, the input image goes through a headless DenseNet-121 network to produce a 3072-dimensional latent vector $\mathbf{z}$. The vector $\mathbf{z}$ is then forwarded to another fully-connected layer (512 units), common to all parameter decoders. From this layer, $5$ output layers are defined: four outputting the $4N$ light source parameters, and one for the ambient term $\mathbf{a}$. Since they are trained separately (sec.~\ref{sec:training}), the $d_i$ layers also take in the position predictions as input.

This network architecture has two key differences compared to the state-of-the-art, non-parametric lighting estimation method of Gardner et al.~\cite{gardner-sigasia-17}. First, while \cite{gardner-sigasia-17} required decoders to produce HDR intensity and RGB from the latent vector $\mathbf{z}$, we instead use simple parametric decoders to directly infer $\mathcal{P}$ from $\mathbf{z}$. Doing so offers the advantage of reducing the number of parameters needed in the decoder, and results in a much faster network. The second key difference is that we leverage a standard feature extractor, DenseNet121 pre-trained on ImageNet, as the encoder. We found this made for more stable learning, avoids a computationally costly pretraining step, and enables us to train solely on the Laval Indoor HDR Dataset (as opposed to \cite{gardner-sigasia-17} which also had to train on SUN360 LDR panoramas~\cite{xiao-cvpr-12}).

\subsection{Training procedure}
\label{sec:training}

While using a parametric modeling of lighting had the advantage of simplifying the neural network architecture, it imposes significant complications on the training process. Indeed, how can we compute a loss on a discrete set of light sources when many of these light sources are not even visible in the image? 

For this, one could dynamically \emph{assign} each predicted light to the closest ground truth light source according, for example, to their angular distance. However, this creates a dependency between the estimated light position $\mathbf{\hat{l}}$ and all the other parameters since the assignment is made based on $\mathbf{\hat{l}}$. This causes problems earlier in the training process, when predicted light directions could change arbitrarily. This creates ``poor'' light assignments that, in turn, lead to unstable gradient flow and network convergence. It also creates ambiguities, such as when two estimated light sources are close to a single ground truth light, etc. We find that training in such a way results in RMSE and si-RMSE several times over the results reported in sec.~\ref{sec:results}, that use the following two-step training procedure.


\subsubsection{Training step 1: radius, color and position}
\label{sec:training-1}

In the first training step, we bypass the need for assigning predicted lights to ground truth light sources by rendering an environment map from the parameters and comparing it to a ground truth environment map:
\begin{equation}
    \mathcal{L}_1(\mathcal{P}, \hat{\mathcal{P}}) = w_{r} \ell_2(f(\hat{\mathcal{P}}), \mathcal{R}) + w_a \ell_2(\hat{\mathbf{a}}, \mathbf{a}) \,,
\end{equation}
where $\mathcal{R}$ is the ground truth environment map associated with $\mathcal{P}$, $f(\cdot)$ is a function mapping $\hat{\mathcal{P}}$ to a environment map, and $\ell_2$ is the L2 loss. To obtain the ground truth environment map $\mathcal{R}$, we use the same warping operator as in \cite{gardner-sigasia-17}, and threshold the warped environment map to 5\% of its peak intensity value, in order to keep only the brightest light sources. The remaining energy is averaged and assigned to the ground truth ``ambient term'', $\mathbf{a}$. Since we care more about detecting the important light sources in the scene, we scale the losses for the light sources and the ambient term differently; we set $w_r=20$ and $w_a=1$.

The differentiable projection function $f(\cdot)$ could be implemented via existing differentiable renderers~\cite{Li:2018:DMC}, however we experimentally found them to be unstable when the parameters are far from their target, in addition to requiring significant computations for each scene. Instead, we implement $f(\cdot)$ by projecting each light source on the sphere onto a spherical gaussian using the mapping
\begin{equation}
    f( \hat{\mathcal{P}}, \mathbf{u}) = \sum_{i=1}^N \hat{\mathbf{c}}_i\exp{\frac{\hat{\mathbf{l}}_i \cdot \mathbf{u} - 1}{\frac{1}{4\pi}\hat{s}_i}} \,,
\end{equation}
where $\mathbf{u}$ is a unit vector giving the direction on the sphere. The angular size $s_i$ is scaled such that the light intensity falls under 10\% of its peak after this distance. $f(\cdot)$ can be computed efficiently since $\hat{\mathbf{l}}_i \cdot \mathbf{u}$ can be precomputed for all possible values of $\hat{\mathbf{l}}_i$. Moreover, $f(\cdot)$ is differentiable everywhere on the sphere, allowing us to backpropagate the error in the environment reconstruction to the predicted 3D light parameters and to the rest of the network. 

Our parametric lighting-to-environment map projection function plays an important role in our method. As noted above, parametric lights give us a compact, low-dimensional representation of scene illumination as well as access to individual lights in the scene. However, an assignment-based optimization approach that relies on correspondences between predicted and ground truth light sources is akin to a greedy, ``local'' optimization scheme which might have bad convergence behaviour. Converting our parametric lights to a environment map allows us to minimize error ``globally'' and leads to significantly better convergence. It also implicitly handles issues such as a mismatch between the number of predicted lights and ground truth lights, thereby making it possible for us to train our network with \emph{any} number of preset light sources. 

Fig.~\ref{fig:parametric} shows an example of an environment map and its equivalent parametric representation using our projection function. The figure also shows how this parametric representation can be projected at different locations in the scene and account for the variations in lighting.

\subsubsection{Training step 2: depth and refinement}
\label{sec:training-2}

Once the network has converged using the first training step, we find that the network predicts good light source positions, $\mathbf{l}_i$, and we switch to independent parametric losses since the assignment problem raised above no longer applies. In this stage we only fine-tune the individual parametric heads and keep the network frozen up to the 512-d feature vector. The position head $\mathbf{l}$ is also frozen to avoid any unwanted feedback. We add a depth estimation head and condition it on both the DenseNet features, $\mathbf{z}$, and the light source position. We use the following loss function: 
\begin{equation}
    \mathcal{L}_2(\mathcal{P}, \hat{\mathcal{P}}) = 
    \ell_2(\mathbf{a}, \hat{\mathbf{a}})
    + \sum_{i=1}^N 
      \ell_2(d_i, \hat{d}_i) 
    + \ell_2(s_i, \hat{s}_i)
    + \ell_2(\mathbf{c}_i, \hat{\mathbf{c}}_i)
\end{equation}
where the L2 losses $\ell_2$ above are computed with respect to the closest light source (angular distance) in the ground truth set $\mathcal{P}$. If a light source has an angular distance greater than $45^\circ$, then no loss is computed as it is deemed too far away from any ground truth light source to be reliable. 


\subsection{Implementation details}

As mentioned, we use a pretrained DenseNet-121 as encoder for our network. Our entire network contains 8.5M parameters, or about $4\times$ fewer than \cite{gardner-sigasia-17}. During the first training step (sec.~\ref{sec:training-1}), the network is trained for 150 epochs, with an early stopping mechanism based on a small validation set. In the second step (sec.~\ref{sec:training-2}), the network is trained for an additional 50 epochs. In both cases, the Adam optimizer is used with a learning rate of $0.001$ (halved each 25 epochs) and $\beta_1 = 0.9$. The first training pass typically takes 12 hours before convergence on a Titan V GPU, while the second finetuning pass completes in a few hours. A batch size of 48 samples was used.

One advantage of our proposed method over previous work is its efficiency in terms of compute requirements. Our parametric representation requires simpler decoders than the ones required to generate a full environment map as in~\cite{gardner-sigasia-17}. As such, our method is much faster than previous work at test time, executing in roughly 51ms per image on a CPU, compared to 127ms for~\cite{gardner-sigasia-17} or 5 minutes for~\cite{karsch-tog-14}. Our method can also be parallelized on GPU, processing 48 images in around 16ms on an nVidia Titan V.

%% file: results.tex
\section{Evaluation}
\label{sec:results}

\begin{figure*}
    \centering
    \footnotesize
    \setlength{\tabcolsep}{2pt}
    \begin{tabular}{cccccc}
     
       \rotatebox{90}{1st percentile} & 
       \includegraphics[width=0.18\linewidth]{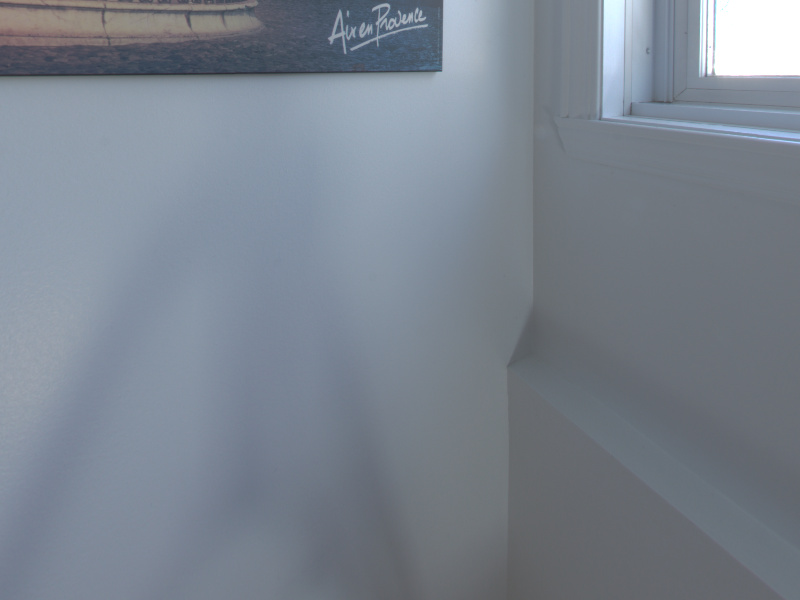} &
       \includegraphics[width=0.2\linewidth]{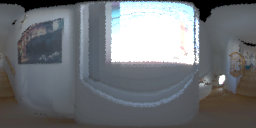} & 
       \includegraphics[width=0.2\linewidth]{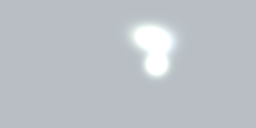} &  
       \includegraphics[width=0.14\linewidth]{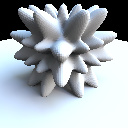} &
       \includegraphics[width=0.14\linewidth]{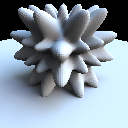} \\
       
       \rotatebox{90}{25th percentile} & 
       \includegraphics[width=0.18\linewidth]{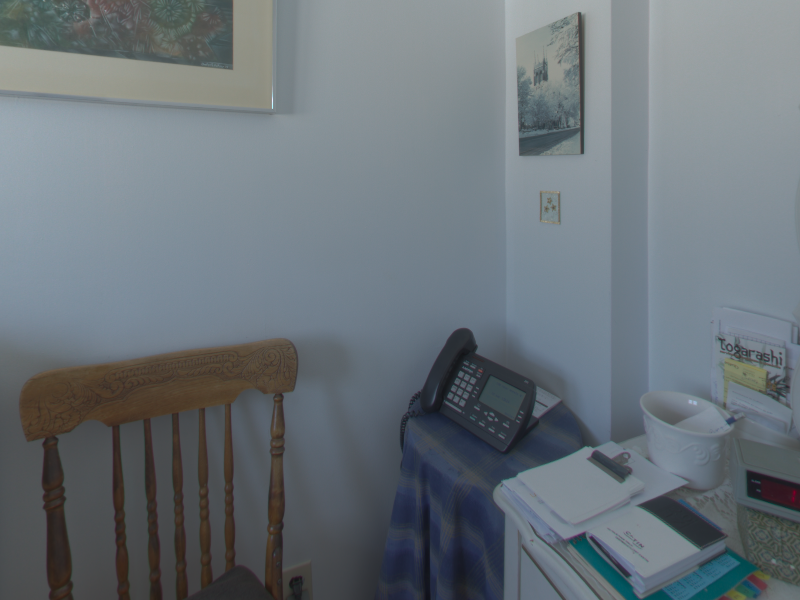} &
       \includegraphics[width=0.2\linewidth]{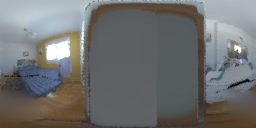} & 
       \includegraphics[width=0.2\linewidth]{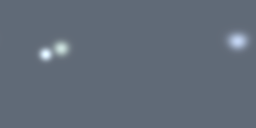} &  
       \includegraphics[width=0.14\linewidth]{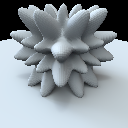} &
       \includegraphics[width=0.14\linewidth]{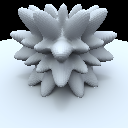} \\
       
       \rotatebox{90}{50th percentile} & 
       \includegraphics[width=0.18\linewidth]{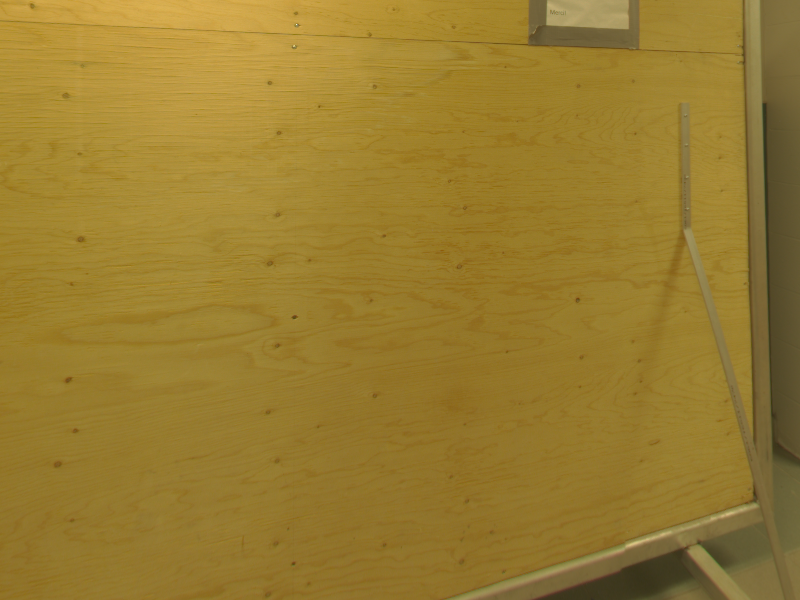} &
       \includegraphics[width=0.2\linewidth]{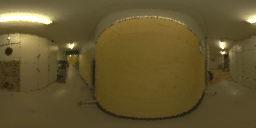} & 
       \includegraphics[width=0.2\linewidth]{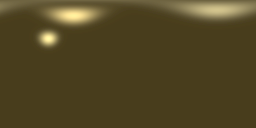} &  
       \includegraphics[width=0.14\linewidth]{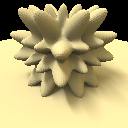} &
       \includegraphics[width=0.14\linewidth]{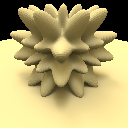} \\
       
       \rotatebox{90}{75th percentile} & 
       \includegraphics[width=0.18\linewidth]{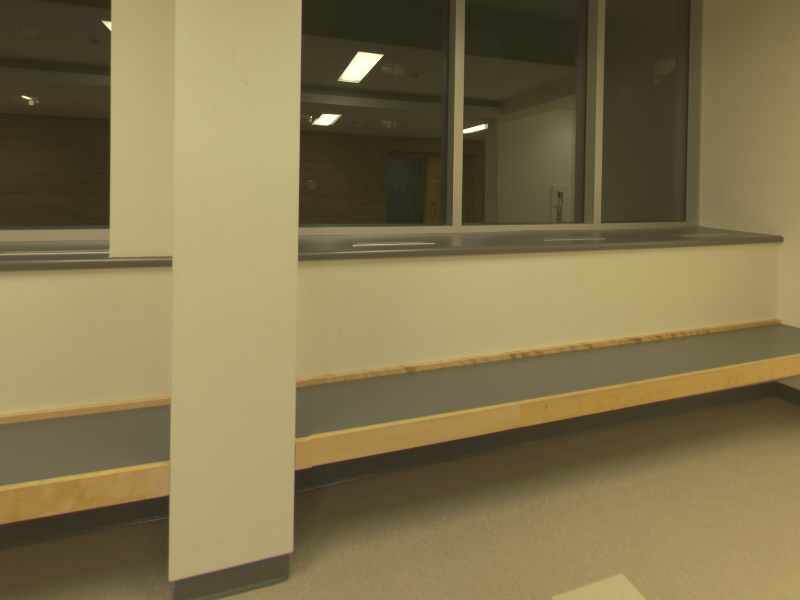} &
       \includegraphics[width=0.2\linewidth]{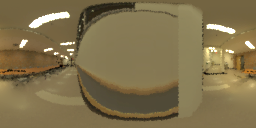} & 
       \includegraphics[width=0.2\linewidth]{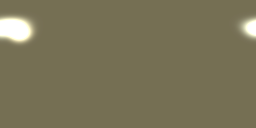} &  
       \includegraphics[width=0.14\linewidth]{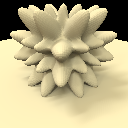} &
       \includegraphics[width=0.14\linewidth]{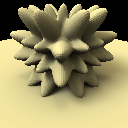} \\

     & (a) Input image & (b) GT EnvyDepth panorama & (c) Our predicted panorama & (d) GT render & (e) Our render \\

    \end{tabular}
    \caption{Examples of errors made by our network, at different RMSE percentiles. The spiky spheres shown on the right are lit by the lighting corresponding to the input image center. Note that even at large error percentiles (bottom), the corresponding renders are plausible.}
    \label{f:percentiles}
\end{figure*}

\begin{figure}
    \centering
    \includegraphics[width=\columnwidth]{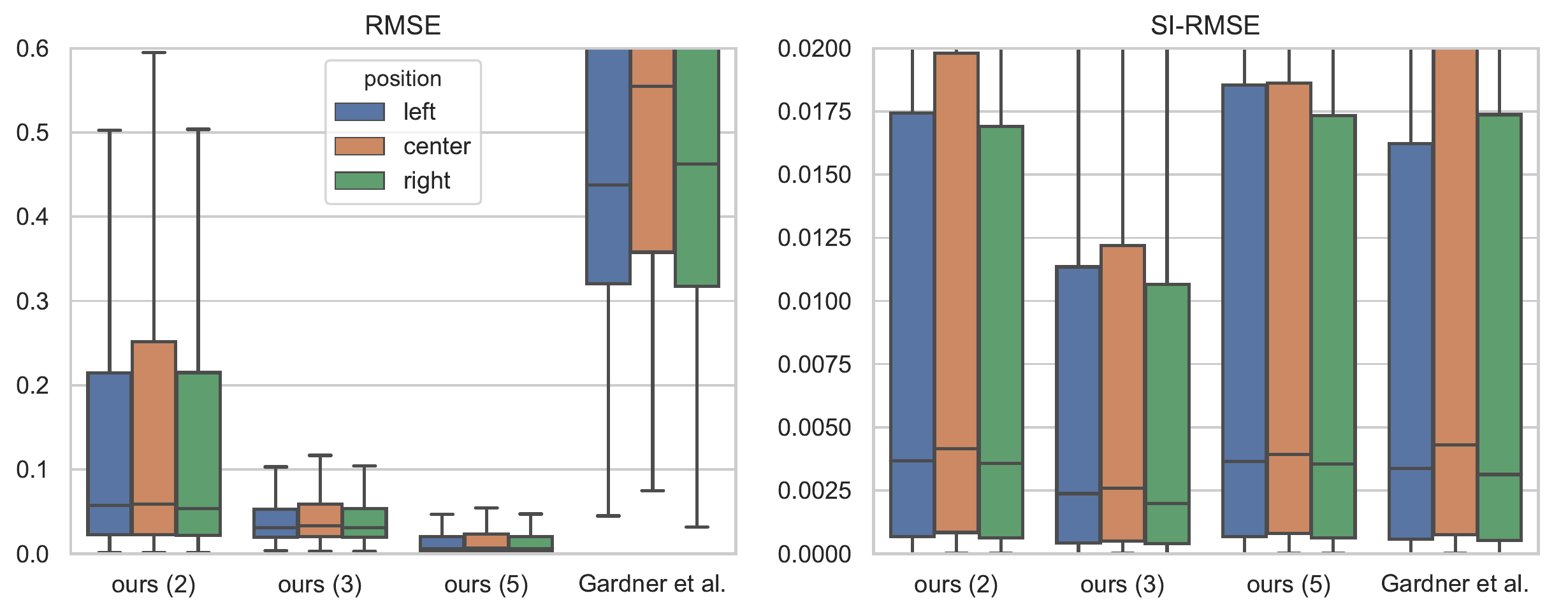}
    \caption{Quantitative comparison between our method configured with 2, 3 and 5 lights and the Gardner et al. method~\cite{gardner-sigasia-17} against the ground truth, as provided by EnvyDepth~\cite{banterle-cgf-13} for (left) RMSE and (right) scale-invariant (si-)RMSE. Results are shown for three insertion position: center and 1m to the left and right. Our method with $N=3$ lights is significantly better in both RMSE and SI-RMSE metrics, compared to~\cite{gardner-sigasia-17}, as asserted by Kruskal-Wallis, Mann-Whitney and Wilcoxon tests with $p<10^{-4}$.}
    \label{f:quantitativecmp}
\vspace{-1.0em}
\end{figure}

Light estimation methods can be tricky to evaluate. In particular, \emph{quantitative} evaluation of realism is not straightforward to define, since realism is in itself ambiguous and mainly relies on human perception. In this section, we present an extensive evaluation protocol which includes both quantitative and qualitative comparisons to the state-of-the-art indoor lighting estimation methods of Gardner et al.~\cite{gardner-sigasia-17} and Karsch et al.~\cite{karsch-tog-14}. We first compute the RMSE between renders made with lighting predictions against a ground truth surrogate (using Envydepth~\cite{banterle-cgf-13}). We then present the results of a user study to quantitatively compare the realism of our results against previous state-of-the-art. 

On the qualitative side, we provide many examples of object insertion (including all 19 user study scenes) on a test set of 252 panoramas from the Laval Indoor HDR Dataset (selected so as to not overlap the training set). We also show virtual object insertion results on stock images (not from our dataset), to demonstrate our approach robustness. Finally, we provide 50 images containing two bunnies, one real and one relit using our prediction. This allows for a direct comparison with an object with a known appearance.

Additionally, our parametric output allows for intuitive user editing, as we demonstrate in supplementary material.



\begin{figure*}
    \centering
    \footnotesize
    \setlength{\tabcolsep}{3pt}
    \begin{tabular}{cccc}
    Ground truth & Ours & Karsch et al.~\cite{karsch-tog-14} & 
    Gardner et al.~\cite{gardner-sigasia-17} \\
    \includegraphics[width=0.22\linewidth]{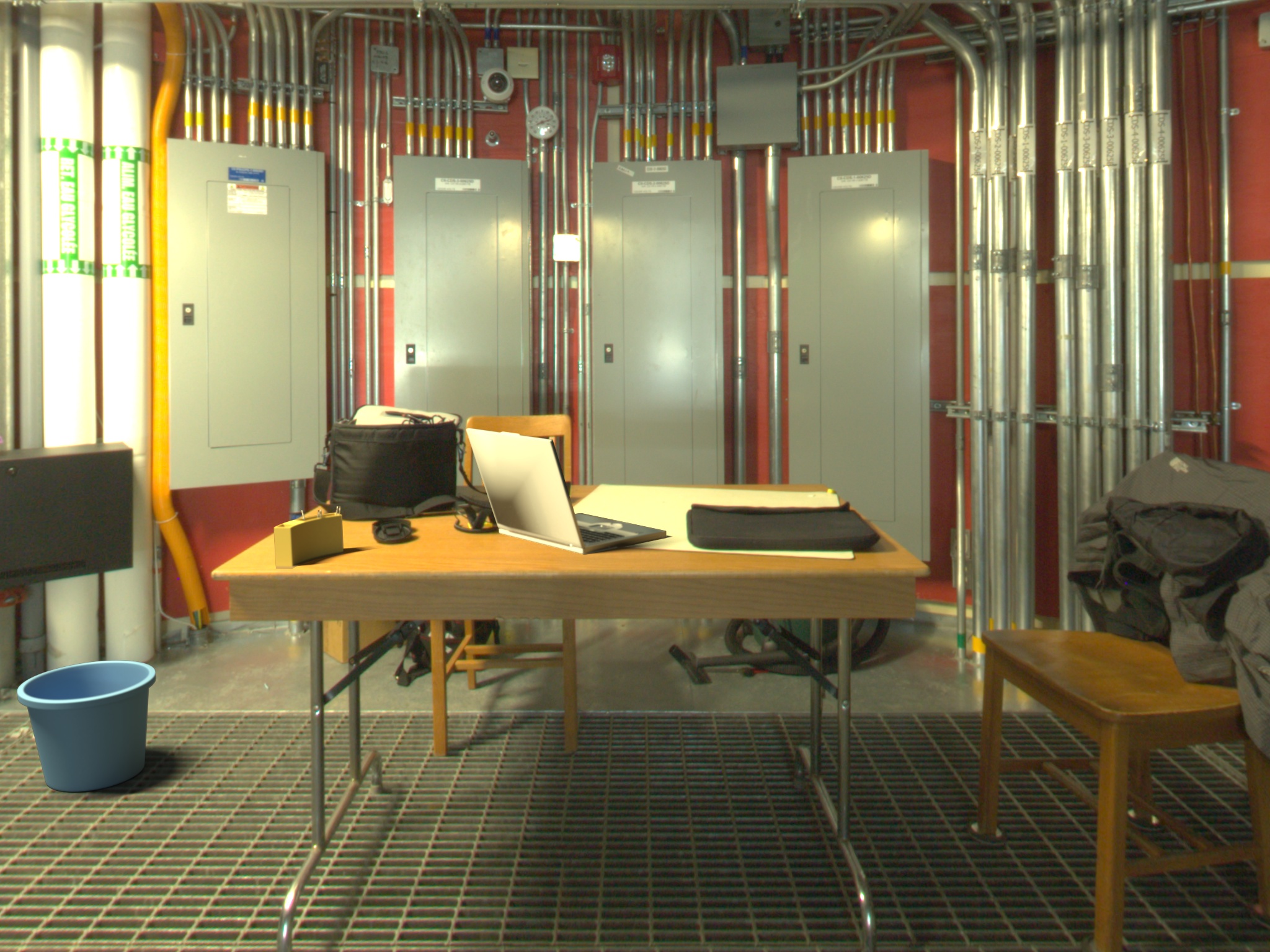} & 
    \includegraphics[width=0.22\linewidth]{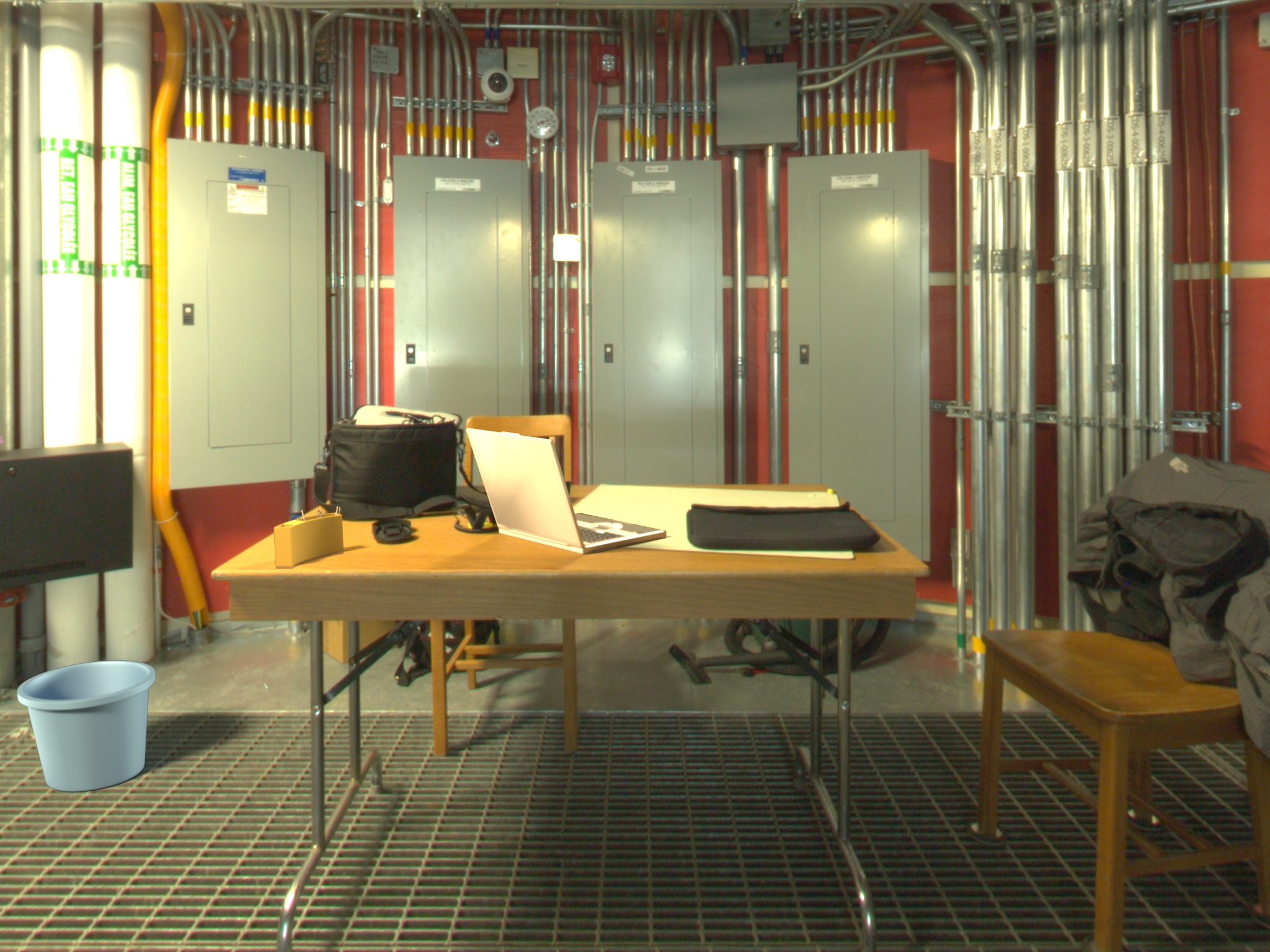} & 
    \includegraphics[width=0.22\linewidth]{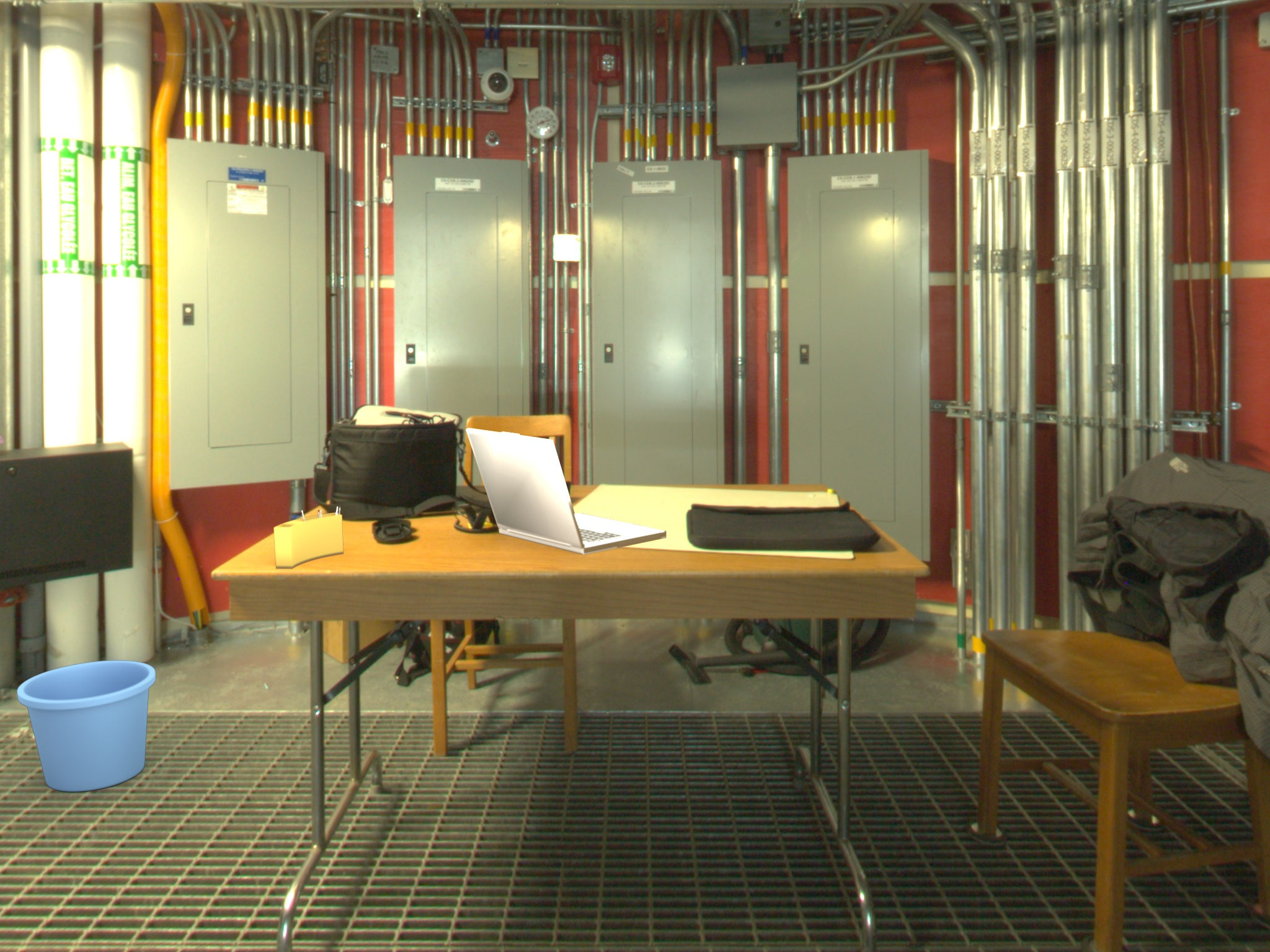} & 
    \includegraphics[width=0.22\linewidth]{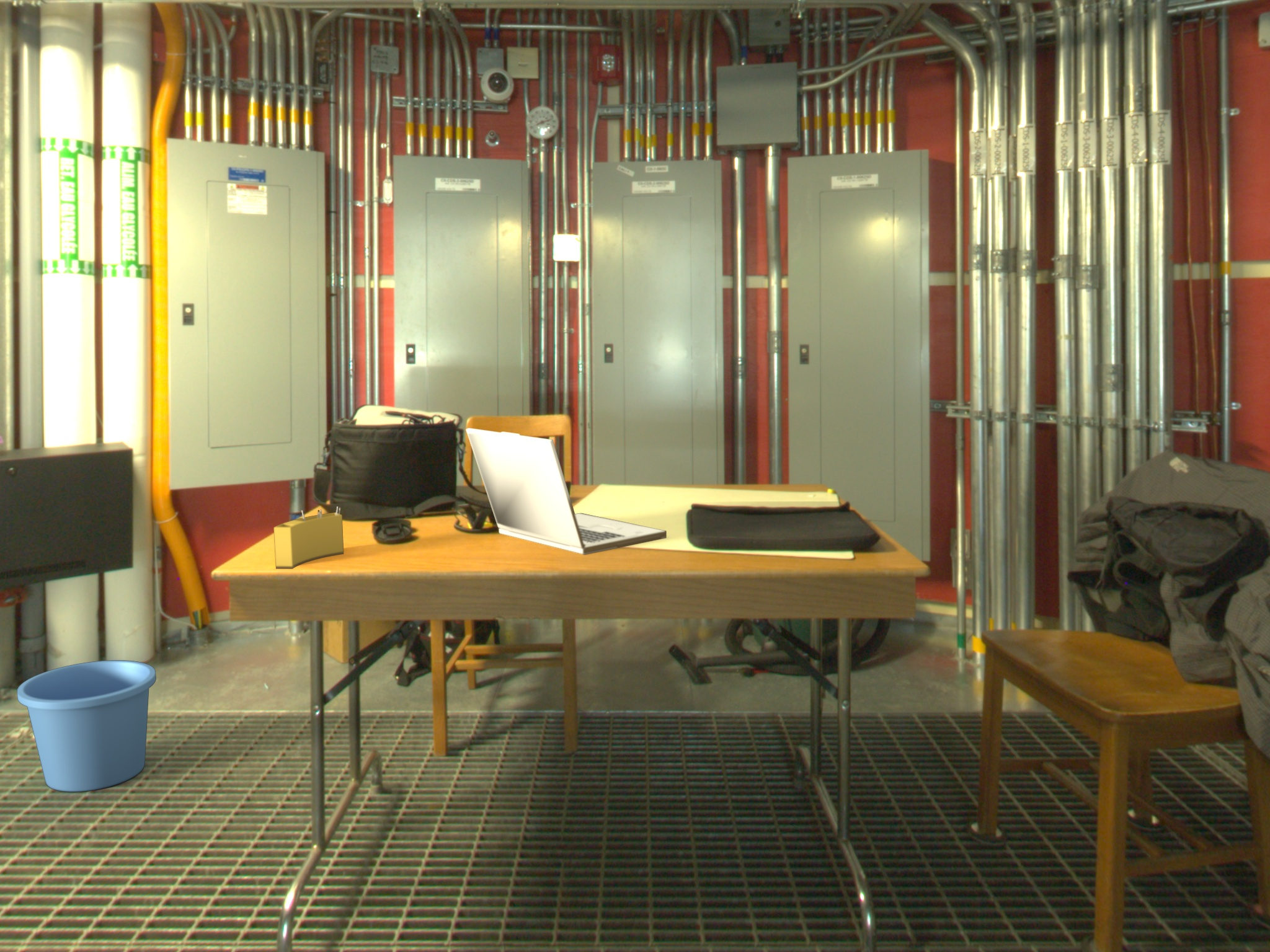} \\ 
    \includegraphics[width=0.22\linewidth]{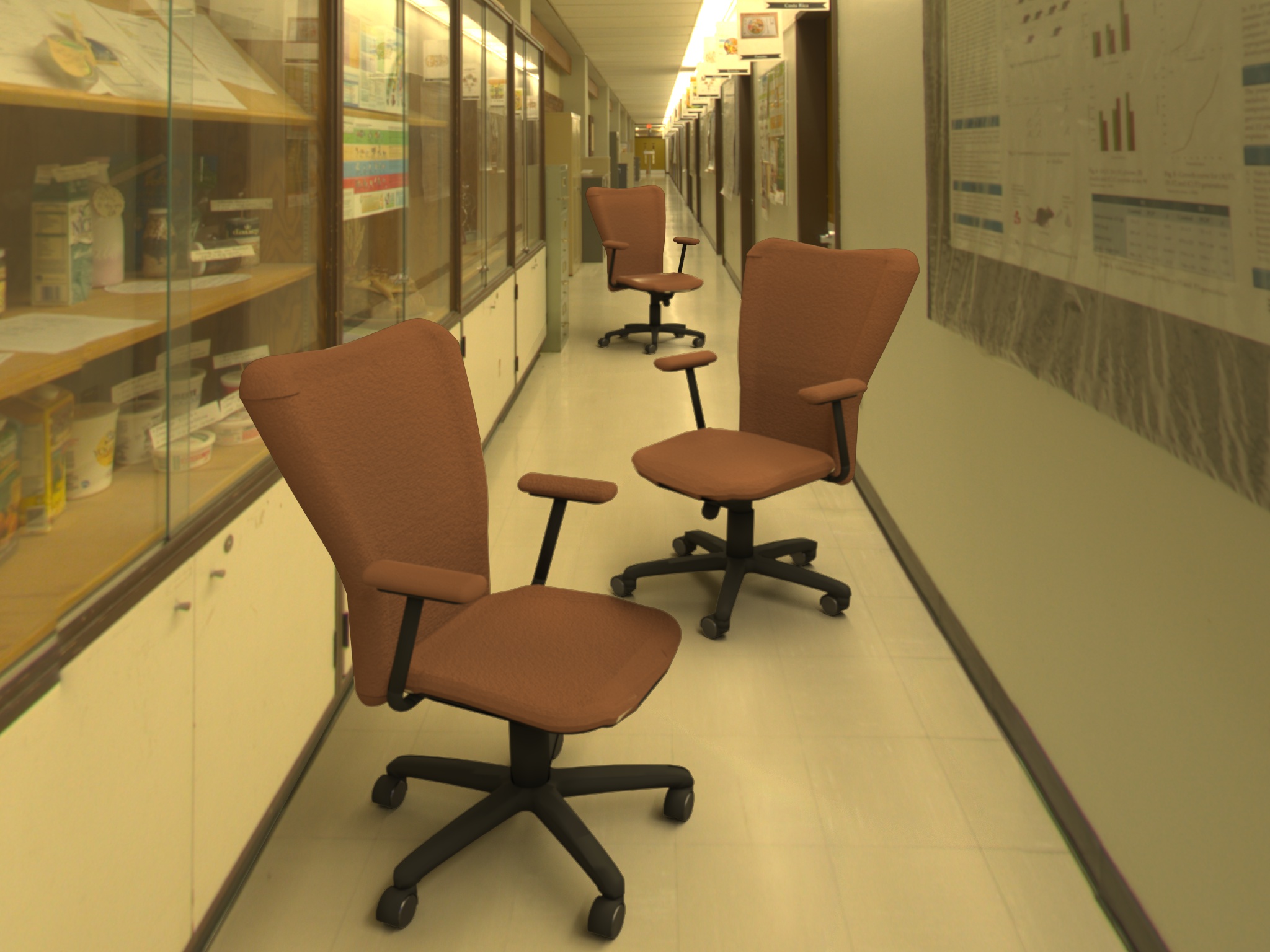} & 
    \includegraphics[width=0.22\linewidth]{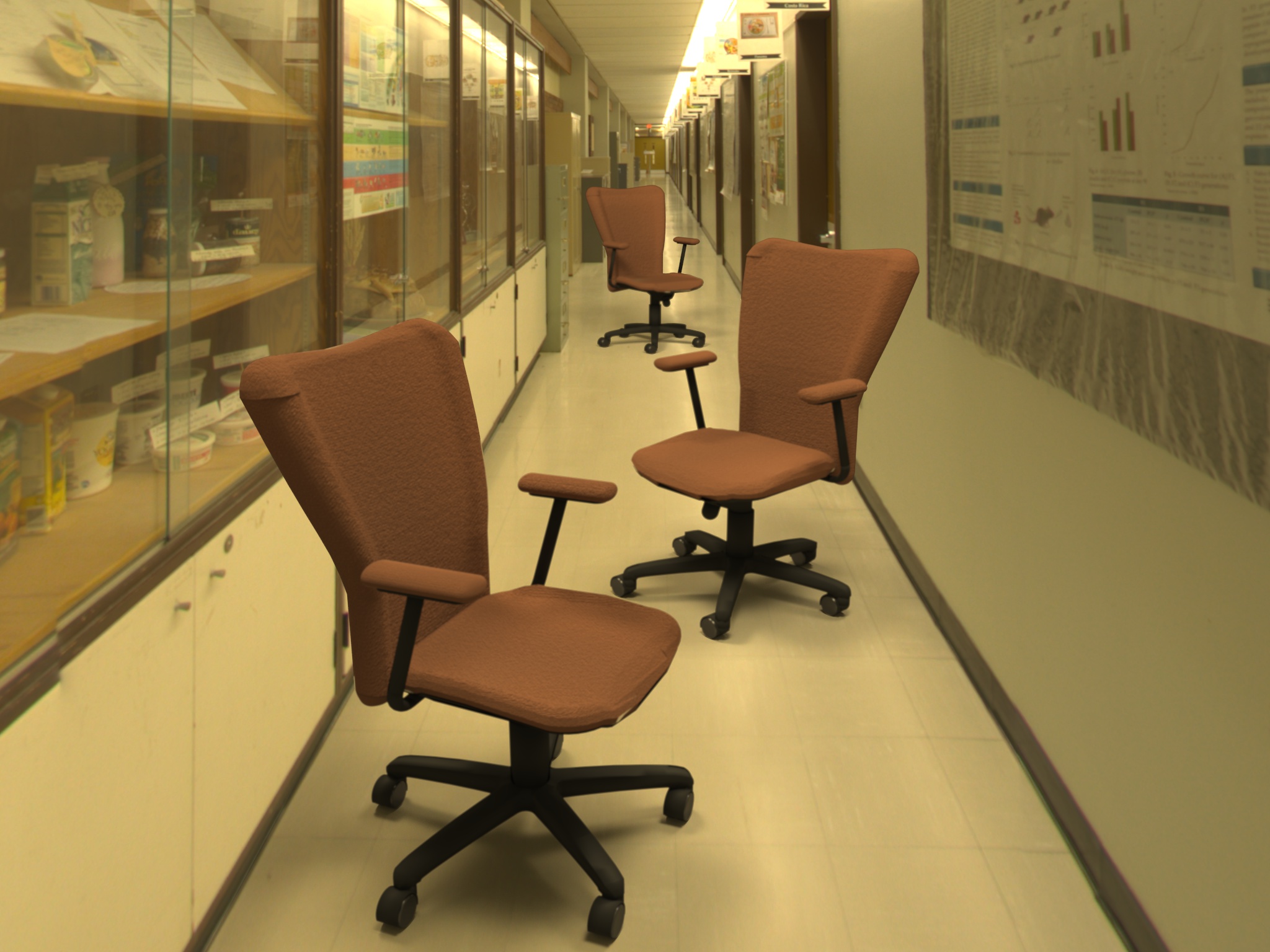} & 
    \includegraphics[width=0.22\linewidth]{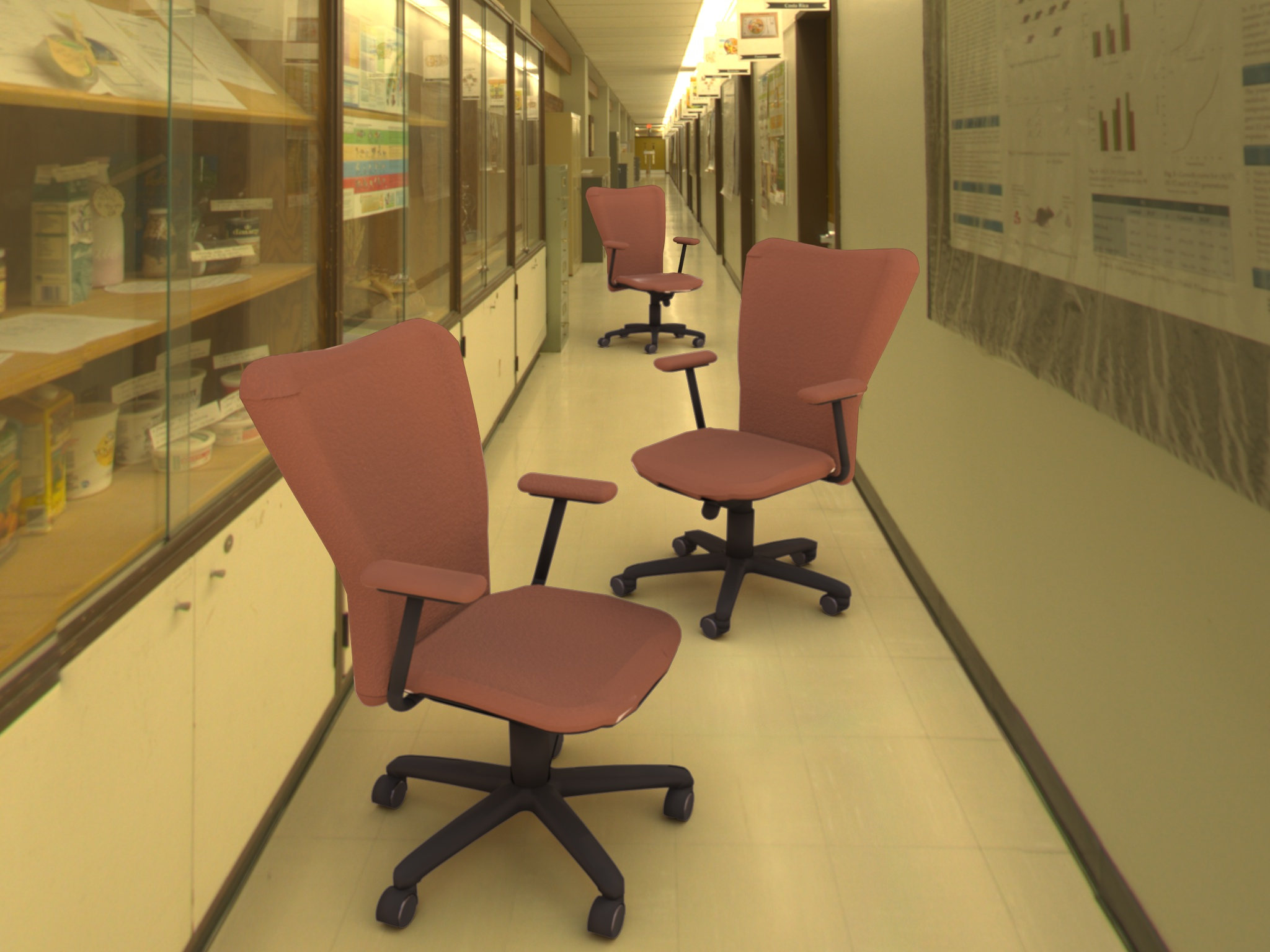} & 
    \includegraphics[width=0.22\linewidth]{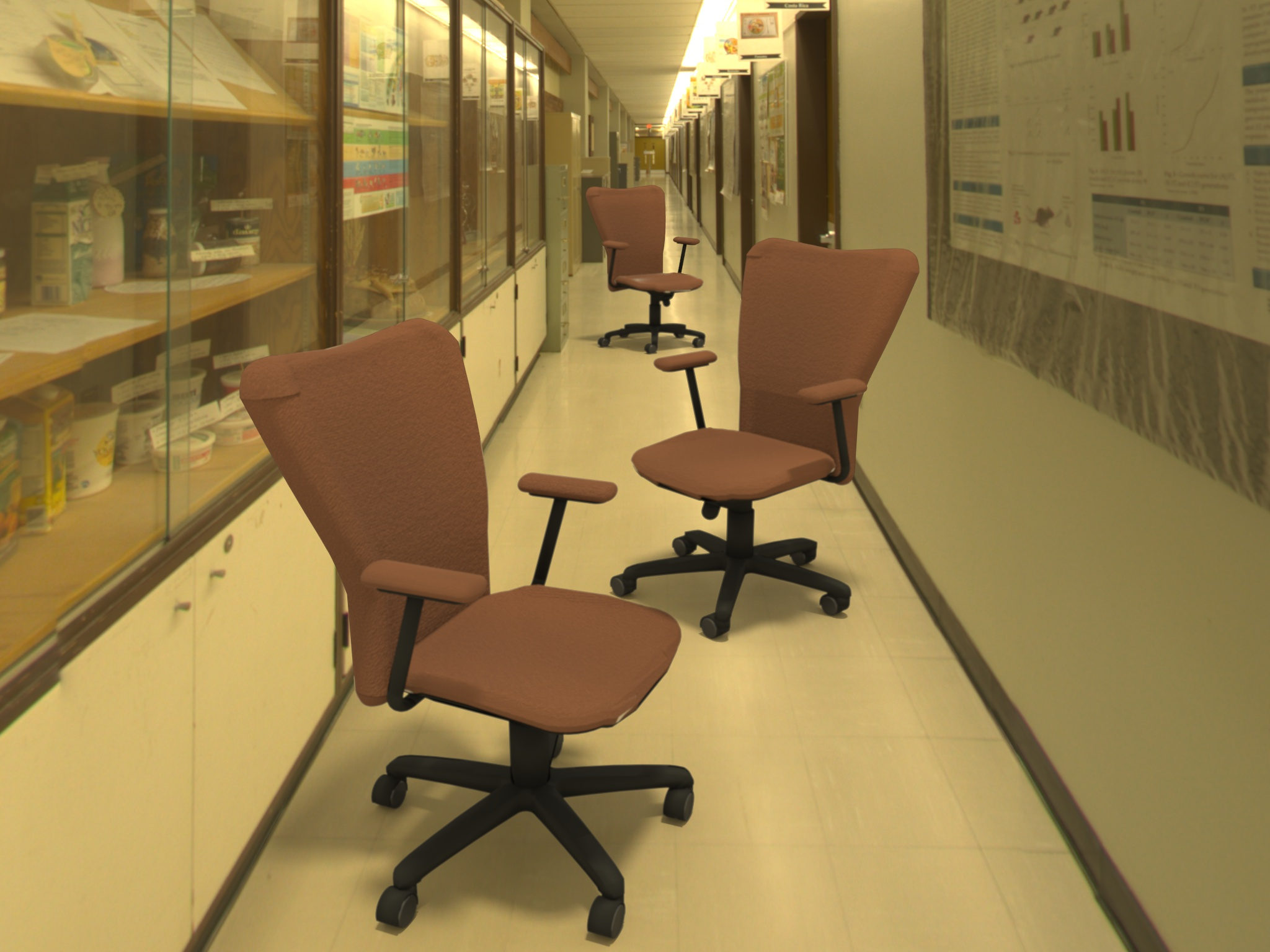} \\ 
    \includegraphics[width=0.22\linewidth]{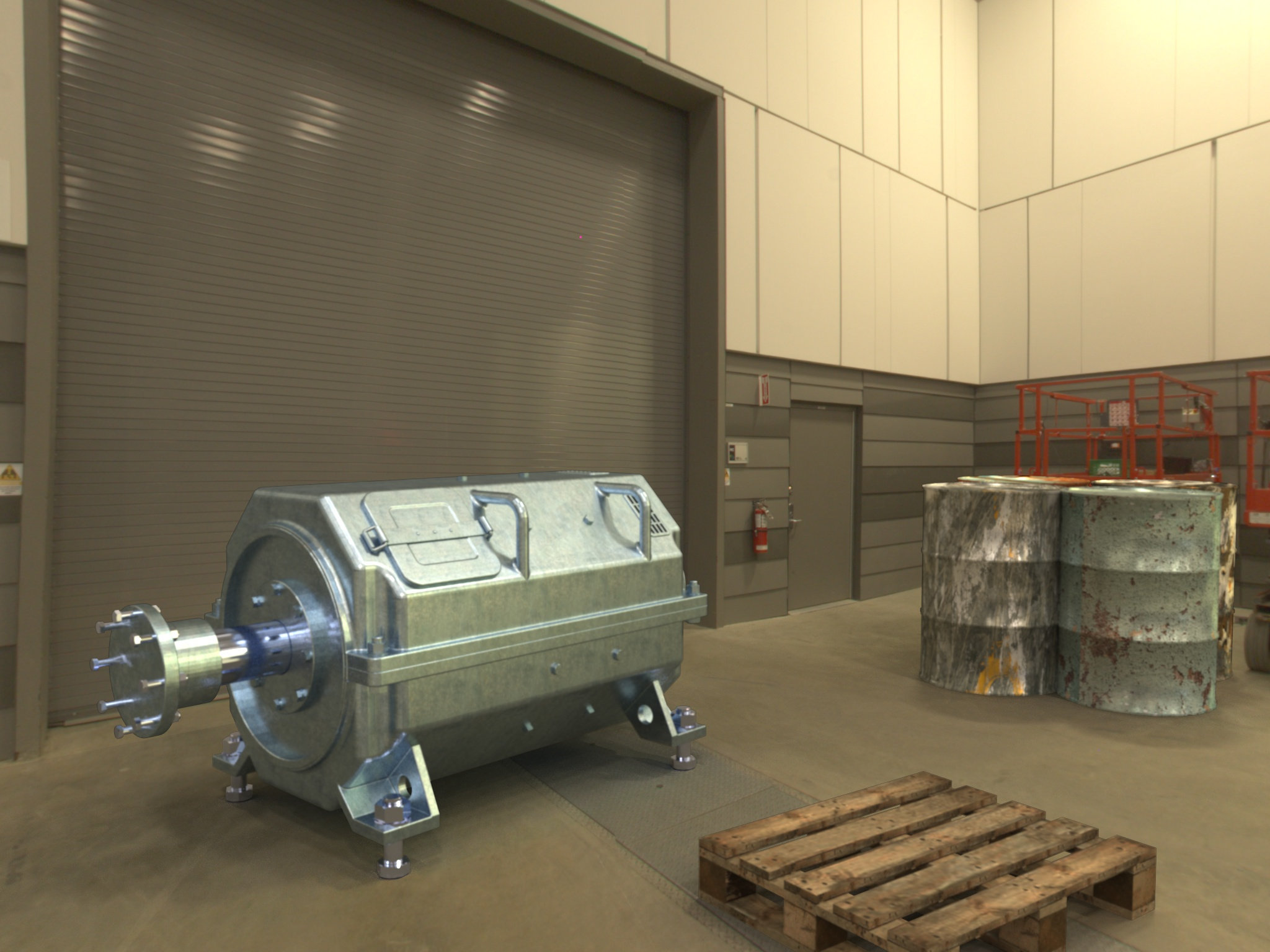} & 
    \includegraphics[width=0.22\linewidth]{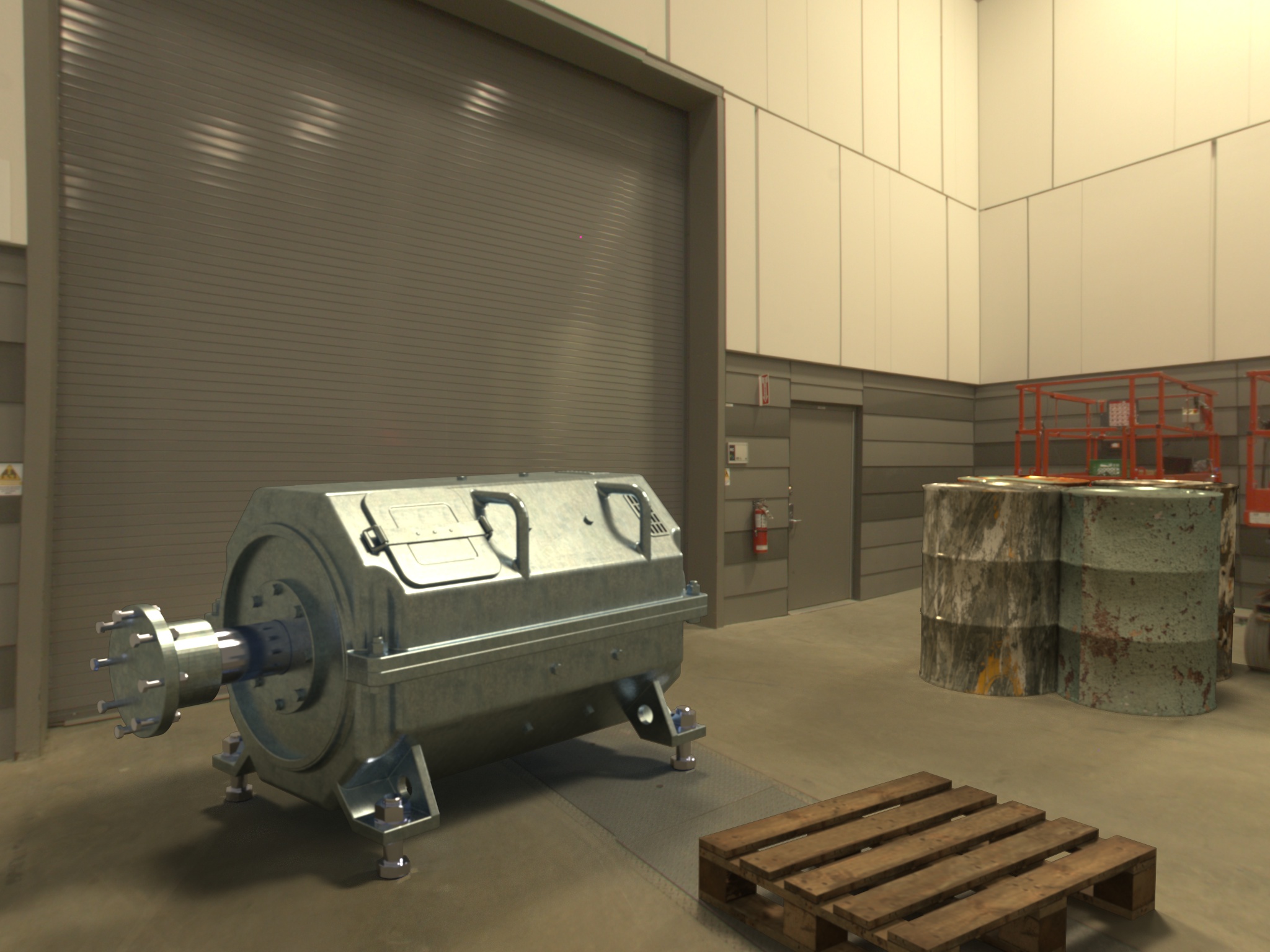} & 
    \includegraphics[width=0.22\linewidth]{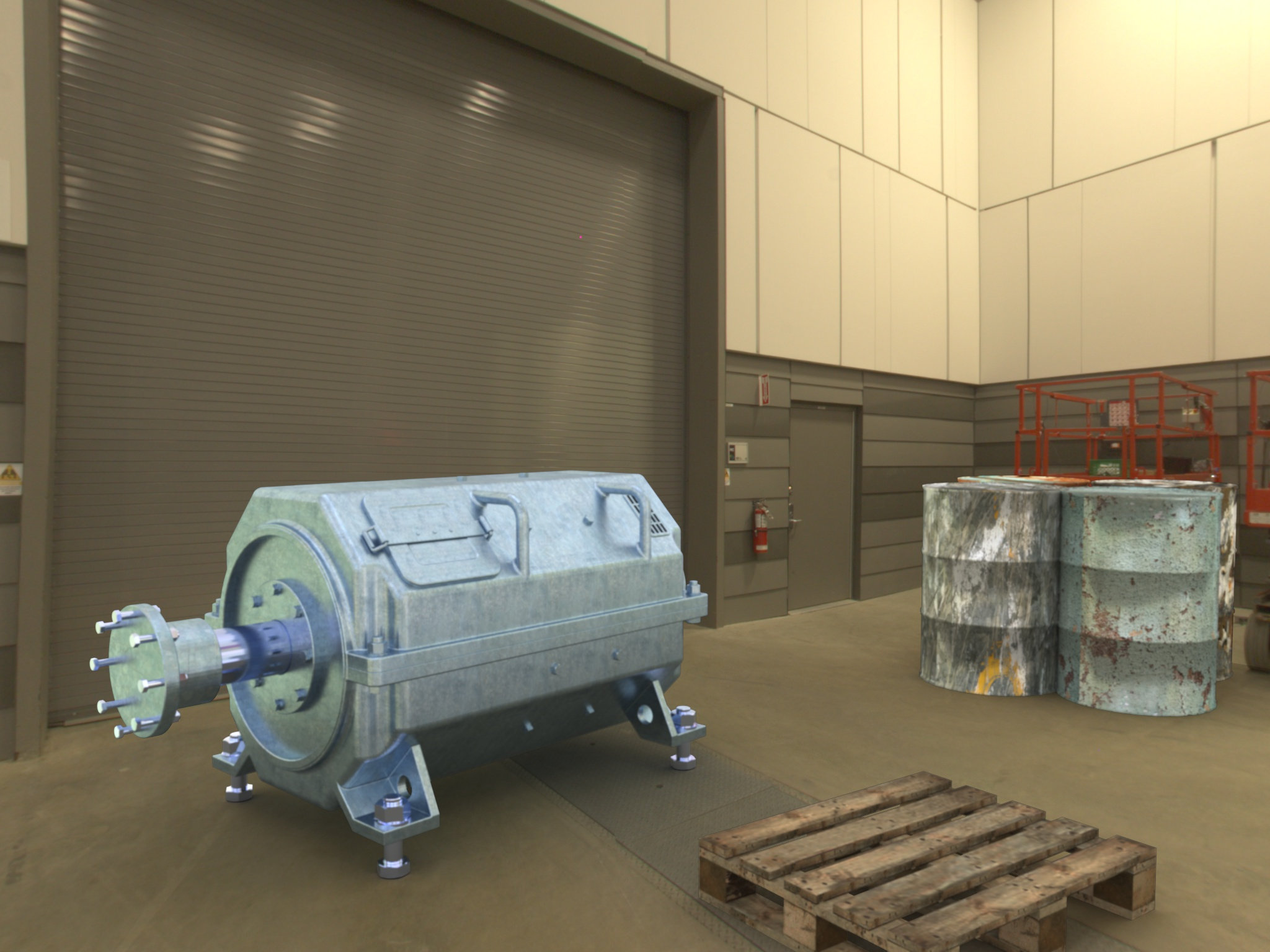} & 
    \includegraphics[width=0.22\linewidth]{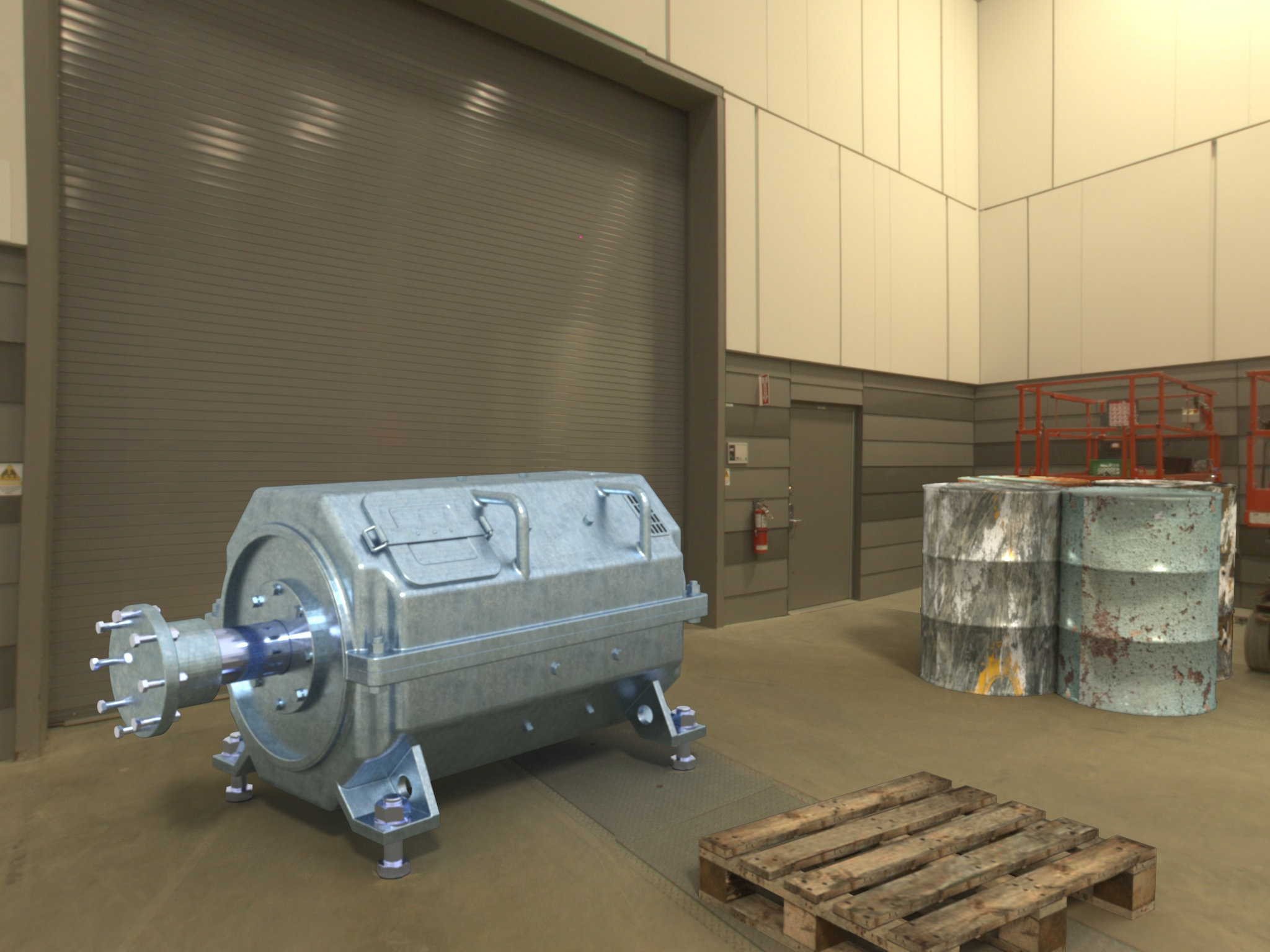} \\*[.25em]
    \end{tabular}
    \caption{Three examples taken from the user study (fig.~\ref{f:userstudy}). From left to right: the ground truth illumination (using ground truth HDR panorama and Envydepth~\cite{banterle-cgf-13}), our proposed method, Karsch et al.~\cite{karsch-tog-14} and Gardner et al.~\cite{gardner-sigasia-17}. Note how our method captures the light color and overall exposure well compared to other methods. All images used in the user study are provided in supplementary material.}
    \label{f:qualitative_userstudy}
    \vspace{-1.4em}
\end{figure*}

\begin{figure}
    \includegraphics[width=\linewidth]{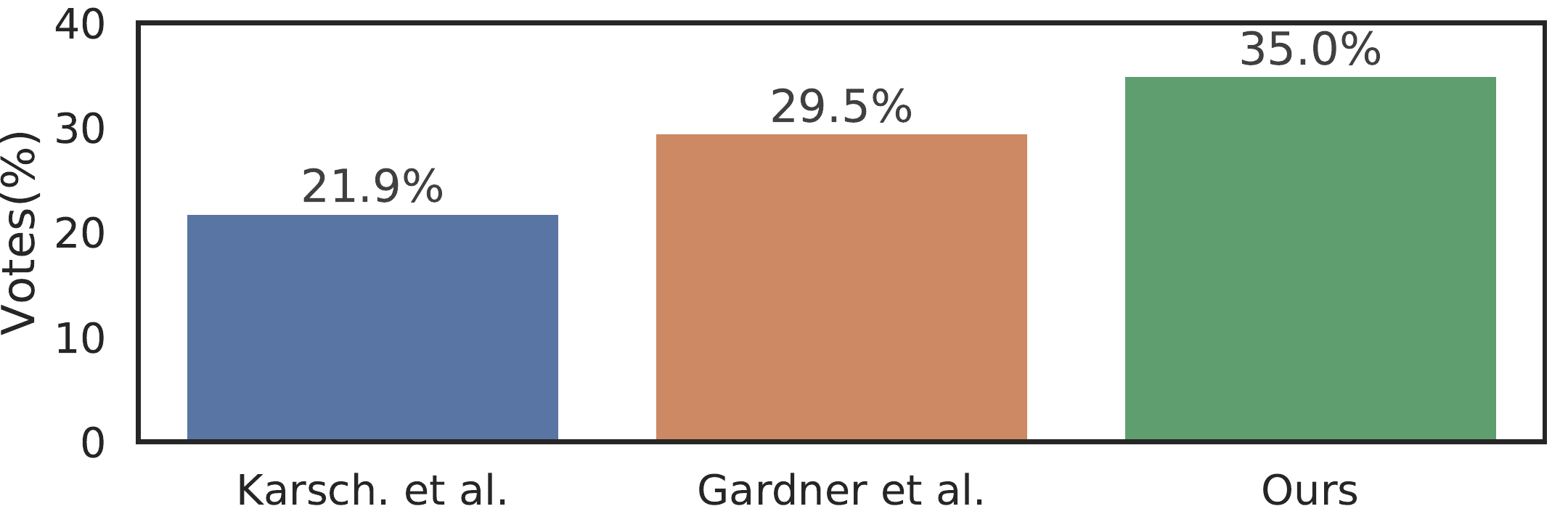}
    \caption{User study results (49 participants, 19 scenes). The percentages denote the fraction of time each method was preferred to the ground truth illumination (perfect confusion = 50\%).}
    \label{f:userstudy}
\vspace{-0.0em}
\end{figure}

\begin{figure}
    \centering
    \hspace{0.2\linewidth}
    \includegraphics[width=0.14\linewidth]{{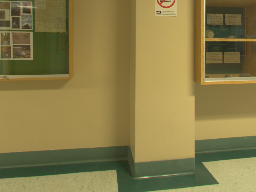}}
    \hspace{0.3\linewidth}
    \includegraphics[width=0.14\linewidth]{{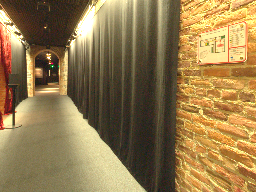}}
    \hspace{0.119\linewidth}
    
    \setlength{\tabcolsep}{1pt}
    \begin{tabular}{ccccc}
     & Env. map & Render & Env. map & Render \\
     (a) &
    \includegraphics[width=0.24\linewidth]{{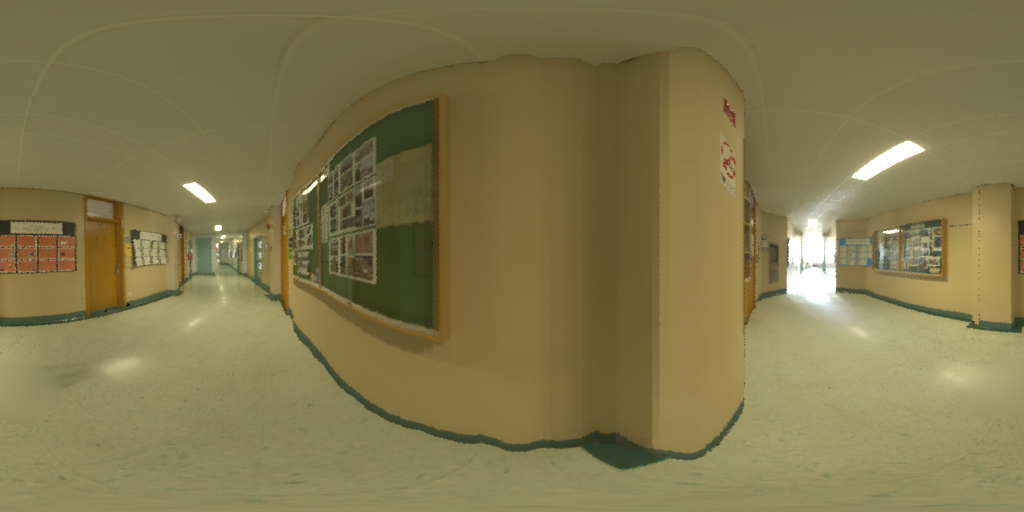}} & 
    \includegraphics[width=0.21\linewidth]{{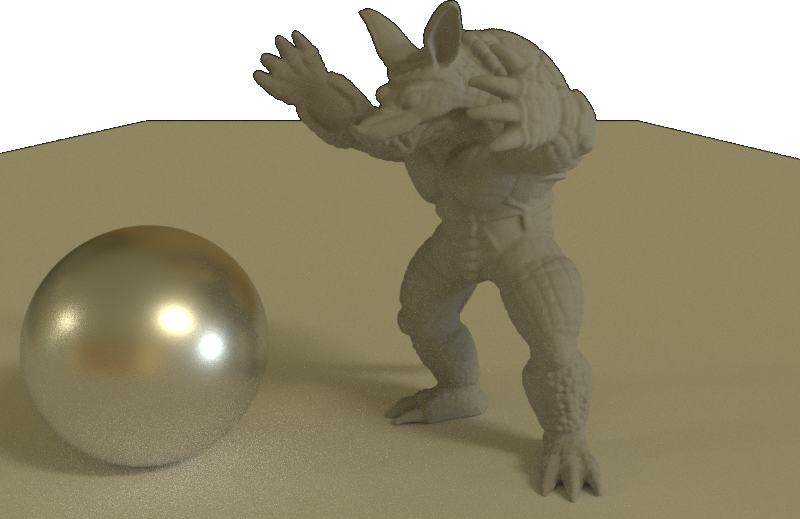}} &
    \includegraphics[width=0.24\linewidth]{{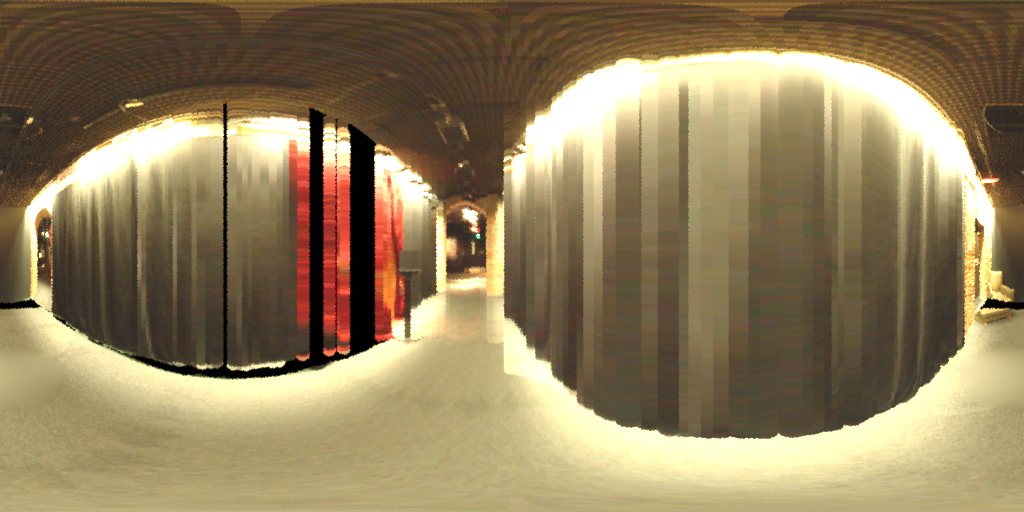}} & 
    \includegraphics[width=0.21\linewidth]{{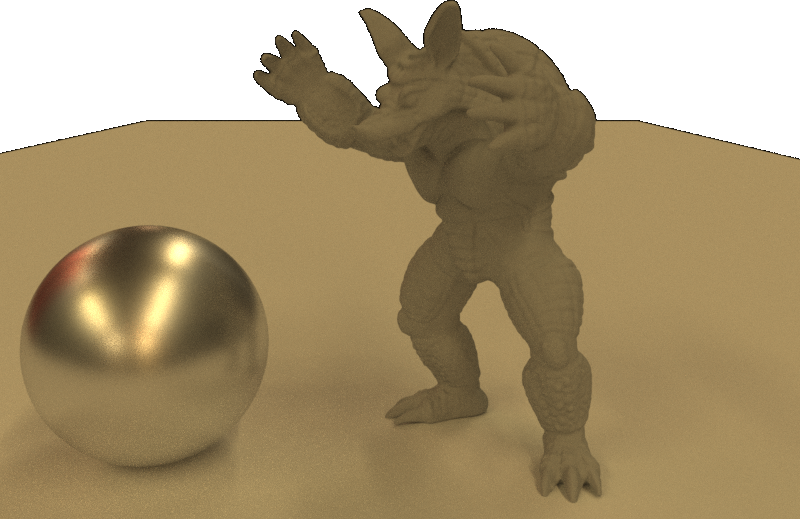}} \\
    
     (b) &
    \includegraphics[width=0.24\linewidth]{{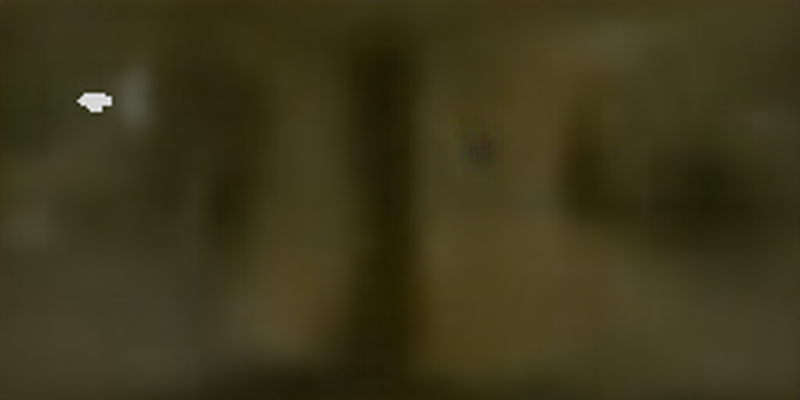}} & 
    \includegraphics[width=0.21\linewidth]{{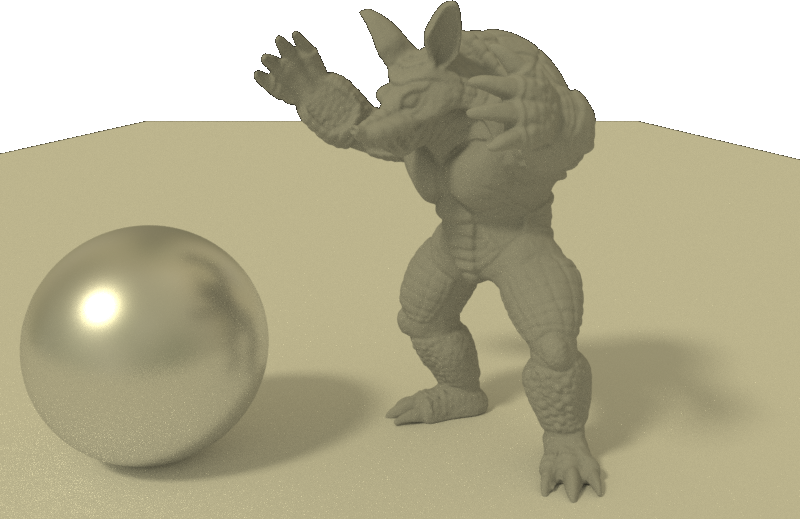}} &
    \includegraphics[width=0.24\linewidth]{{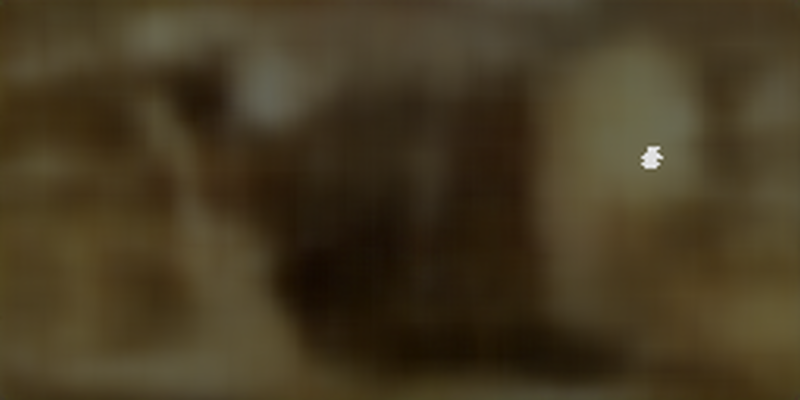}} & 
    \includegraphics[width=0.21\linewidth]{{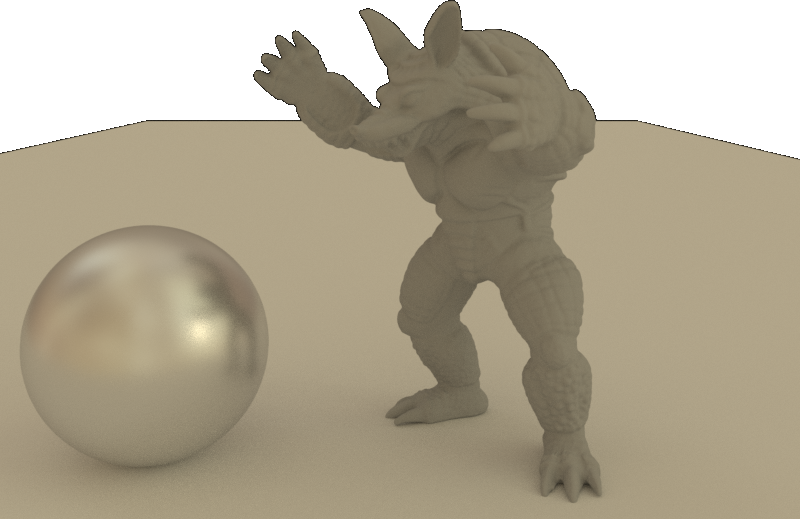}} \\
    
     (c) &
    \includegraphics[width=0.24\linewidth]{{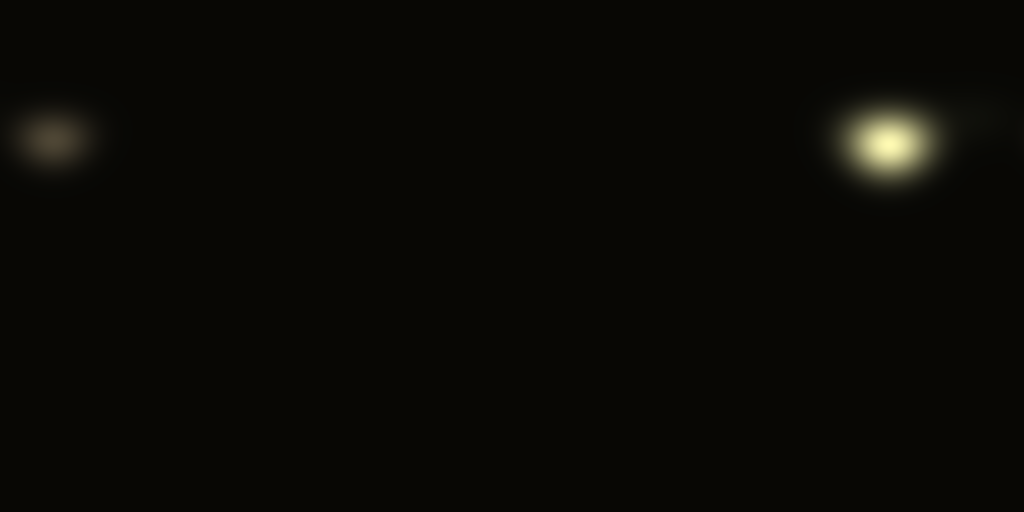}} & 
    \includegraphics[width=0.21\linewidth]{{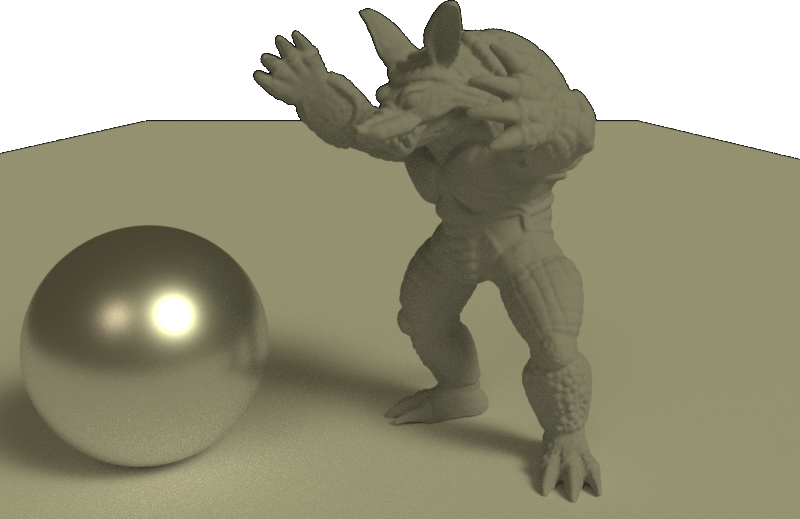}} &
    \includegraphics[width=0.24\linewidth]{{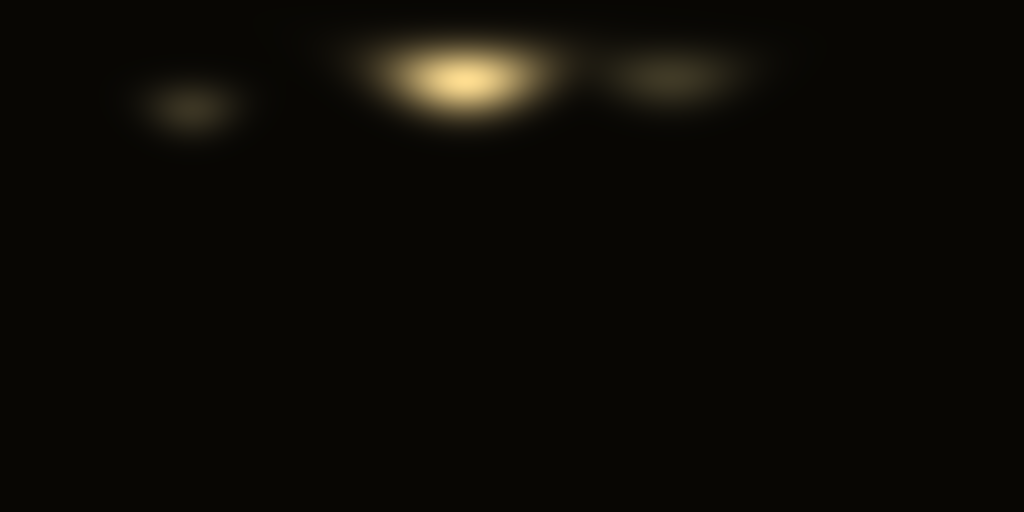}} & 
    \includegraphics[width=0.21\linewidth]{{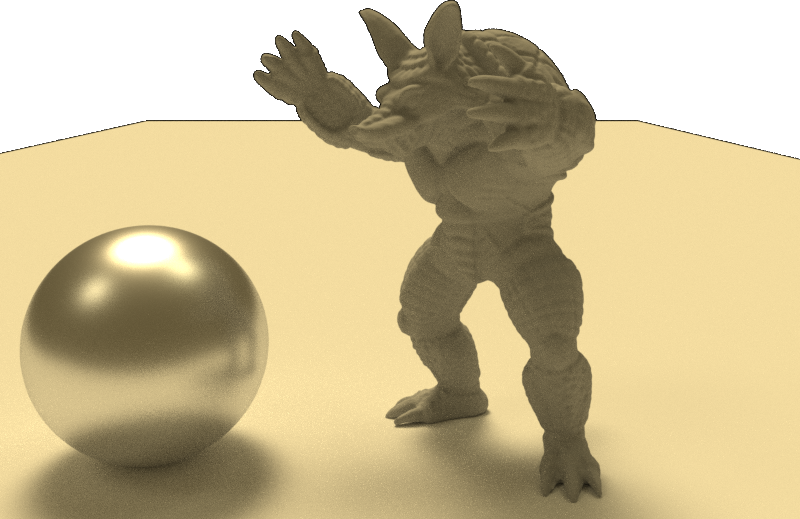}} \\
    \end{tabular}
    
    \caption{Single image lighting estimation results comparing between parametric and non-parametric lighting. We compare environment maps and renders between a) the ground truth illumination, b) the non-parametric approach of Gardner et al.~\cite{gardner-sigasia-17}, and c) our parametric lighting representation. More results in supplementary material.}
    \label{f:qualitativevsgt}
\vspace{-3.2em}
\end{figure}

\subsection{Quantitative evaluation}

We quantitatively compare three different configurations of our method to Gardner et al.~\cite{gardner-sigasia-17} on our test set. To do so, we render a diffuse virtual object---the spiky sphere shown on the right of fig.~\ref{f:percentiles} that captures both shading and shadowing---at three different locations in the scene: in the center, 1 meter to the left, and 1 meter to the right. We do this for each method and the ground truth environment map, warped using the EnvyDepth geometry. 

We evaluate the accuracy of each method by computing the RMSE and scale-invariant (si-RMSE)~\cite{grosse-iccv-09} metrics on the corresponding renderings; these are shown in fig.~\ref{f:quantitativecmp}. RMSE is most sensitive to errors in overall light intensity. We observe that our technique significantly outperforms~\cite{gardner-sigasia-17} in this regard. Furthermore, we see a large improvement in performance as we increase the amount of lights estimated from 2 to 3, and finally 5. This result suggests that adding more lights gives more degrees of freedom to the network to better fit the overall energy distribution. Another improvement of our approach is that our parametrization allows for each light source to have a different color, while \cite{gardner-sigasia-17} only predicts the intensity.
Finally, our errors are consistent across different insertion points indicating our ability to adapt to scene location, unlike \cite{gardner-sigasia-17} which exhibits higher variance. 

The si-RMSE measure factors out scale differences between the different methods and ground truth, and focuses on cues such as shading and shadows, mainly due to light position. Our results show that performance increases from a 2 lights to a 3 lights configuration. However, performance slightly decreases from 3 to 5 lights, suggesting that optimizing the positions of 5 lights is harder than 3 lights. Our method with the 3-lights configuration obtains a 40\% increase in median estimation performance on si-RMSE over \cite{gardner-sigasia-17}. Finally, we note that a network trained using a direct loss from the start (e.g. skipping the first training step, sec.~\ref{sec:training-1}) obtains si-RMSE errors an order of magnitude higher than with our two steps approach, effectively validating the light assignation issues reported in sec.~\ref{sec:training}.
 Fig.~\ref{f:nbrlights} shows a qualitative visualization of the environment maps recovered with a varying number of lights. 

We also compare the methods' realism using a user study. Users were shown pairs of images with composited objects---rendered either with ground truth lighting or a randomly selected method's prediction---and asked to pick the more realistic image. We used 19 scenes with multiple virtual objects scattered in them. Results (49 participants) are shown in fig.~\ref{f:userstudy} and confirm the improvement over~\cite{gardner-sigasia-17} and~\cite{karsch-tog-14}. Examples of these scenes are shown in fig.~\ref{f:qualitative_userstudy}.

\begin{figure}
\centering
    \setlength{\tabcolsep}{4pt}
	\begin{tabular}{cc}
	\includegraphics[width=0.49\linewidth]{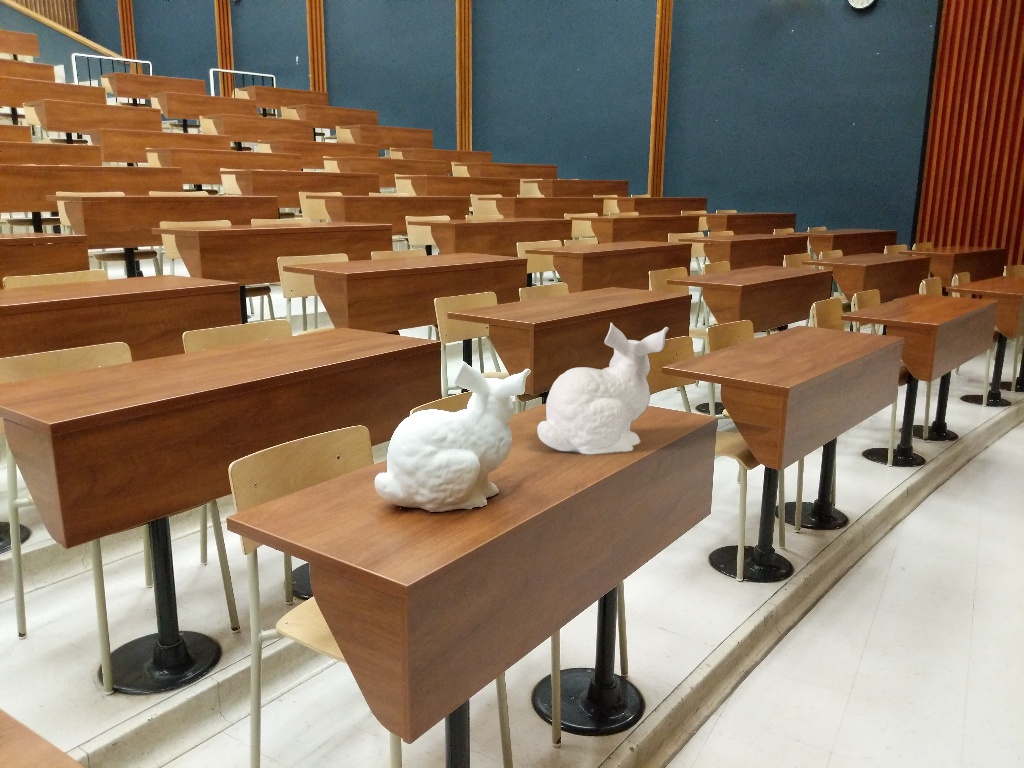}  &
	\includegraphics[width=0.49\linewidth]{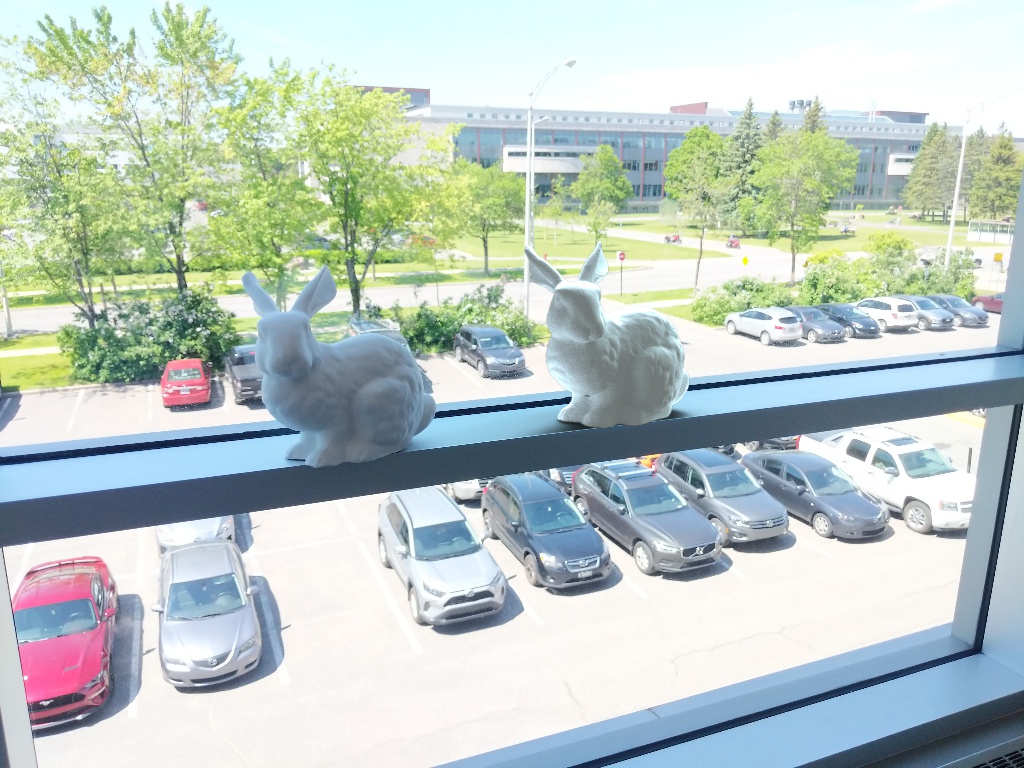} \\
	\includegraphics[width=0.49\linewidth]{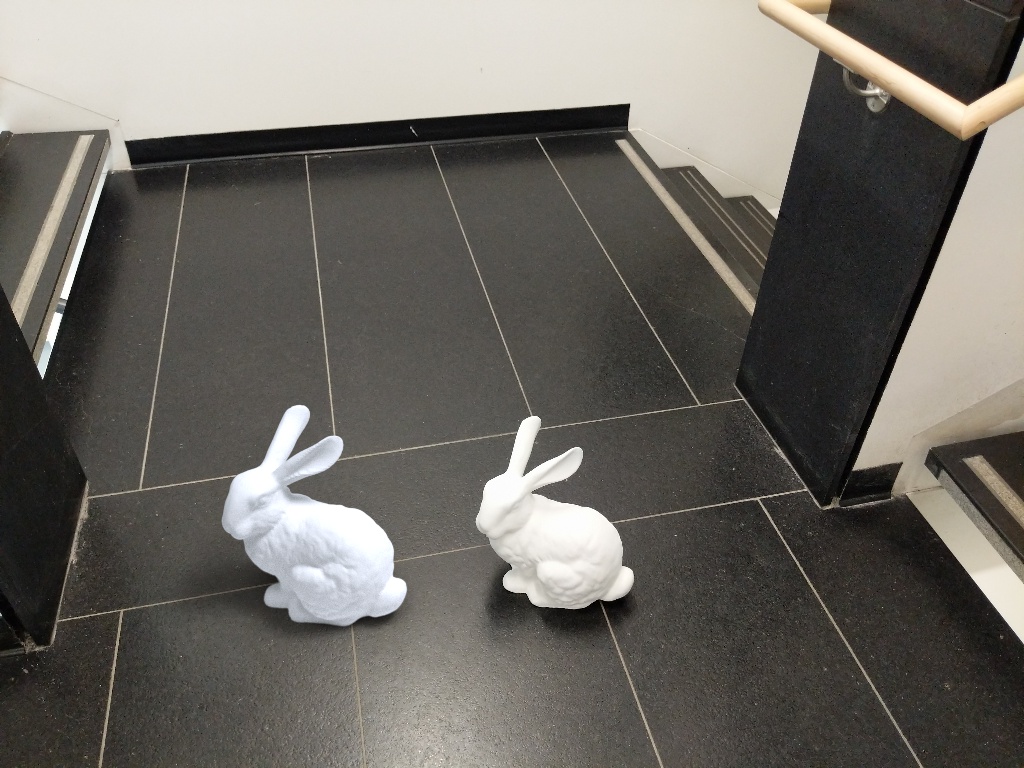} &
	\includegraphics[width=0.49\linewidth]{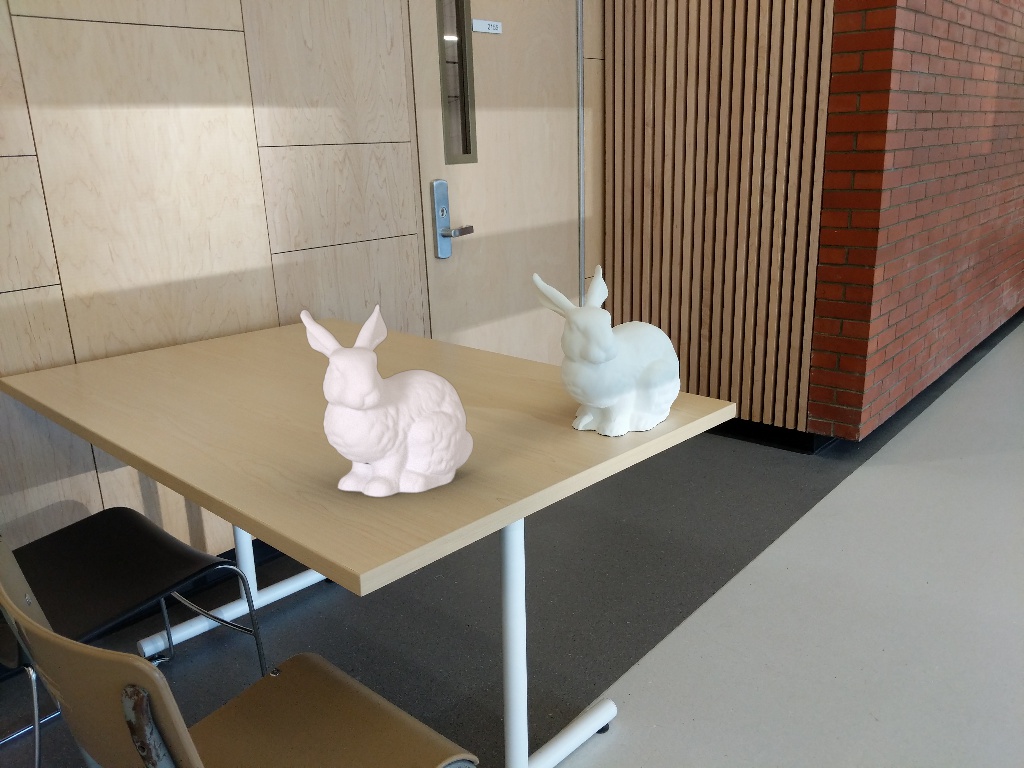} \\
	\includegraphics[width=0.49\linewidth]{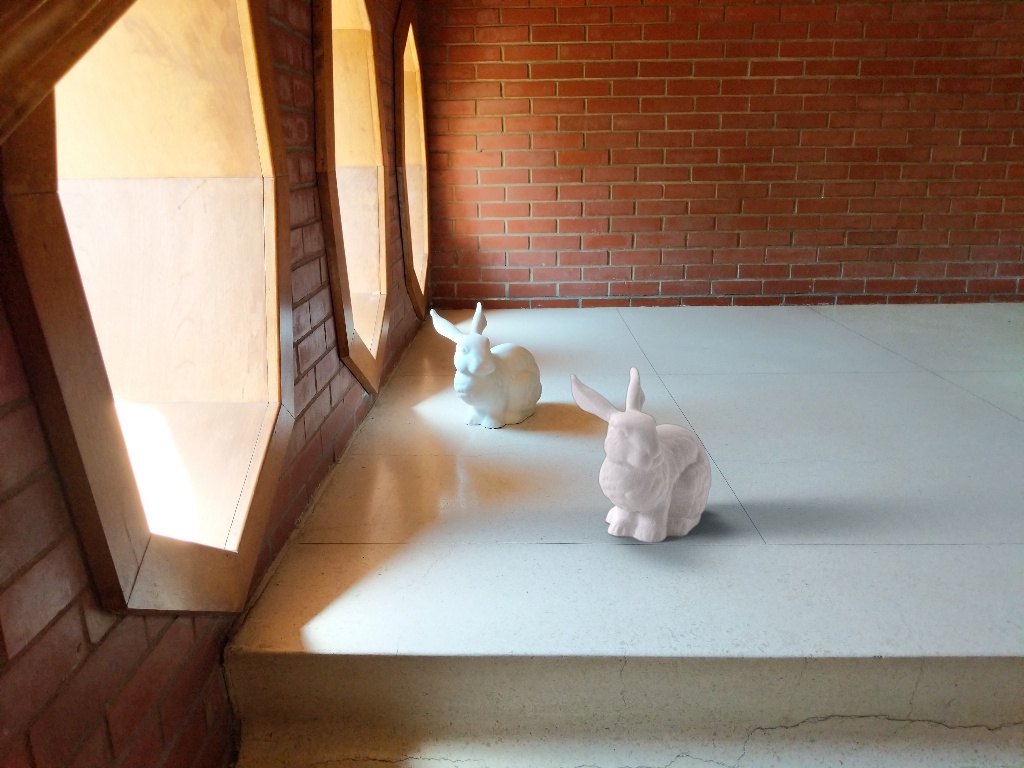} &
	\includegraphics[width=0.49\linewidth]{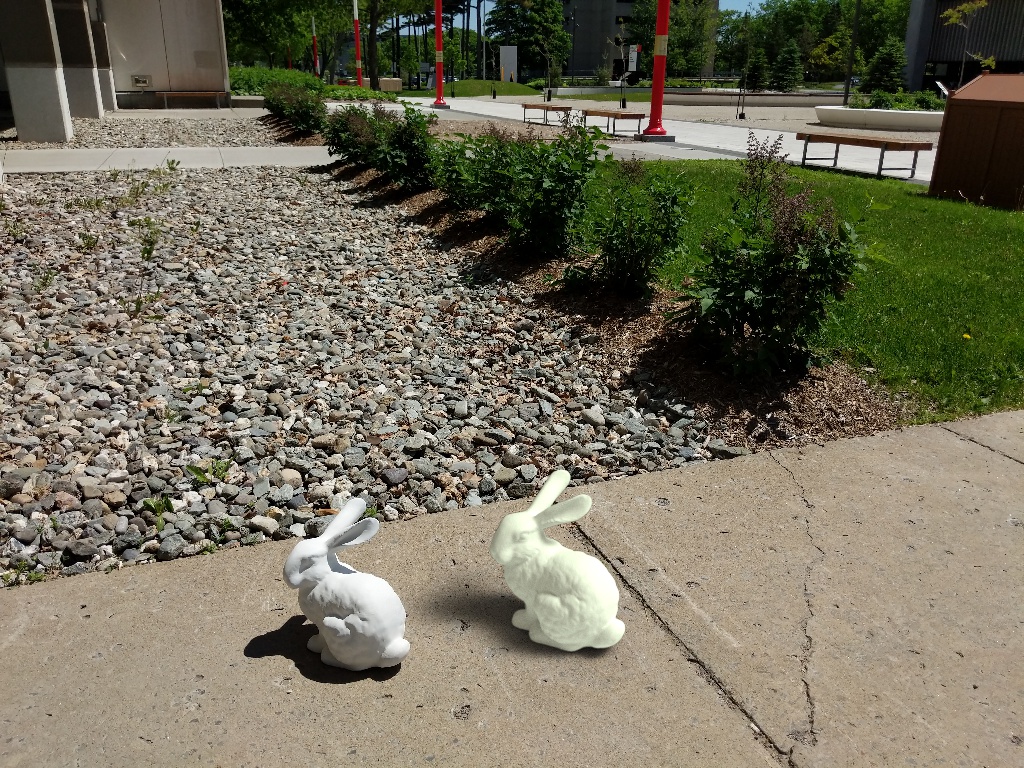} \\
	\end{tabular}
	\caption{Relighting results comparing virtual and real bunnies. Images were captured using a smartphone (Pixel 2, HDR off) in various environments, including outdoors. Our approach can closely match the true lighting, even in situations it was not trained for like outdoor daylight, and reliably produces realistic results.}
	\label{f:bunnies}
\end{figure}

\begin{figure}
    \setlength{\tabcolsep}{1pt}
	\begin{tabular}{cc}
	\includegraphics[width=0.49\linewidth]{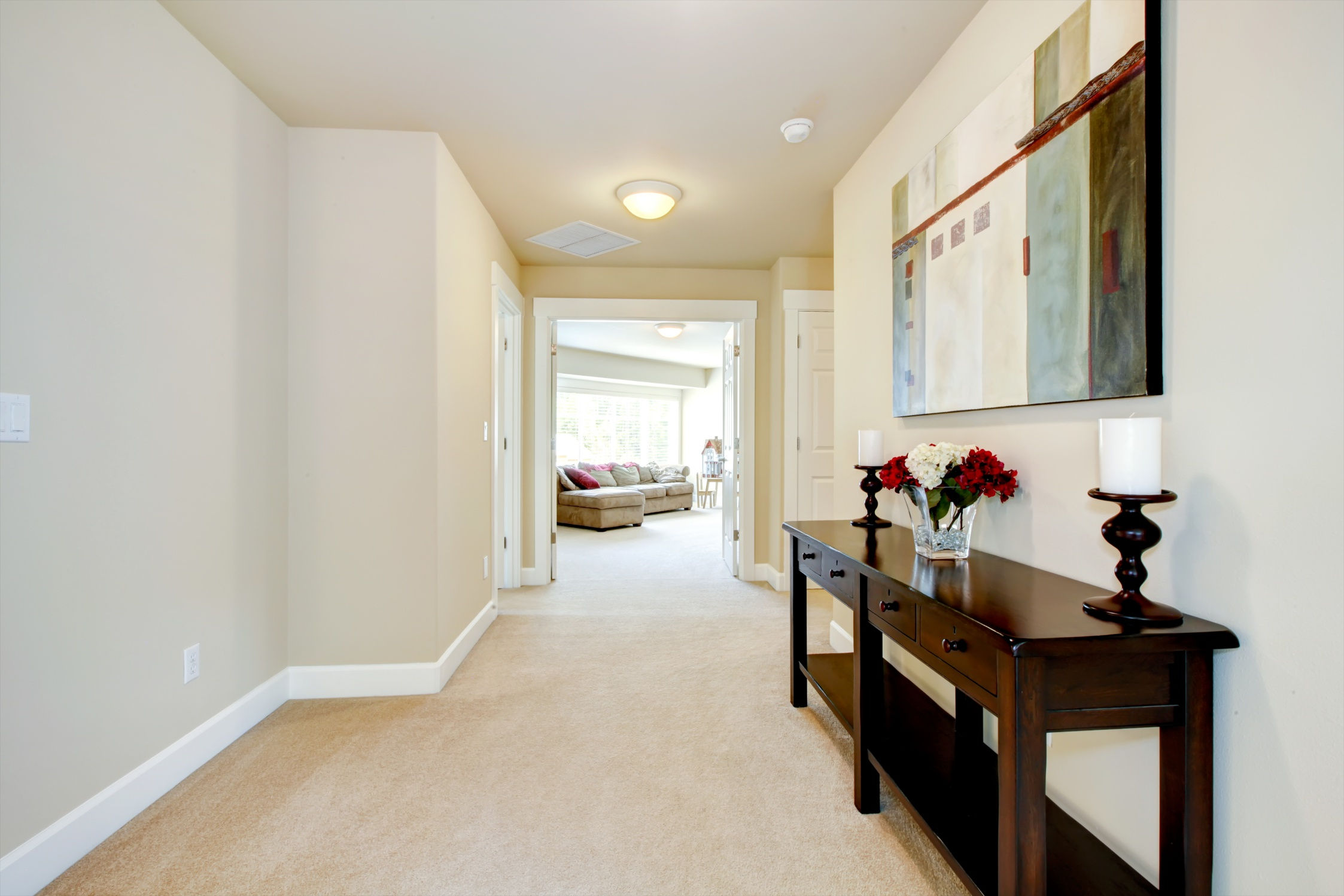}  &
	\includegraphics[width=0.49\linewidth]{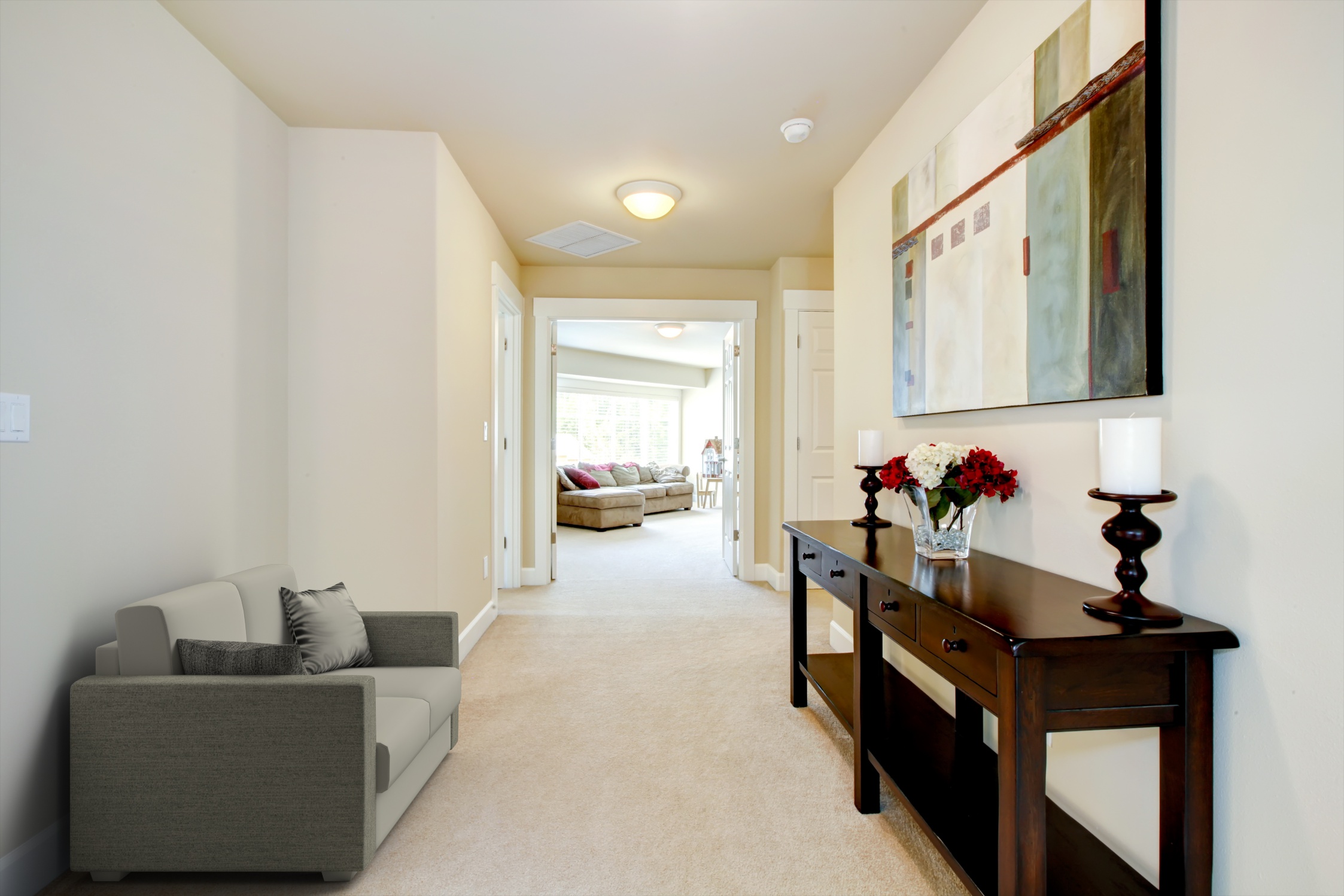} \\
	\includegraphics[width=0.49\linewidth]{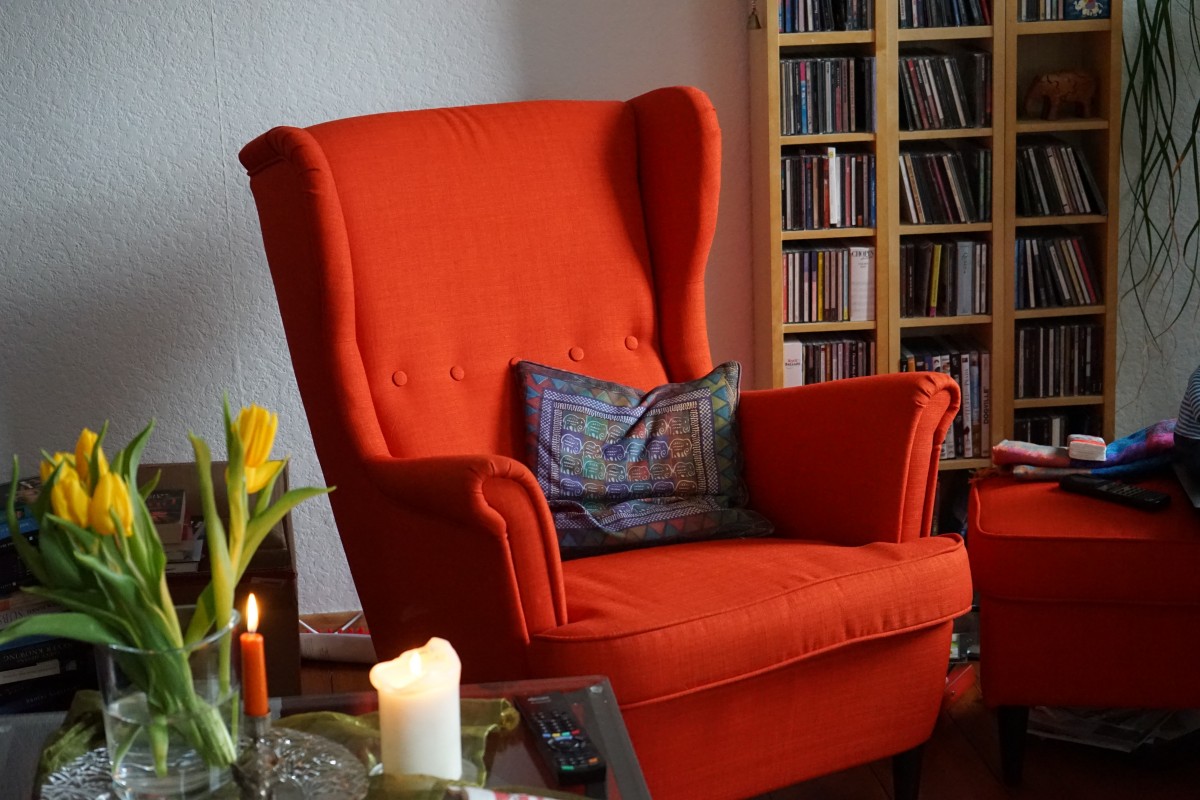} &
	\includegraphics[width=0.49\linewidth]{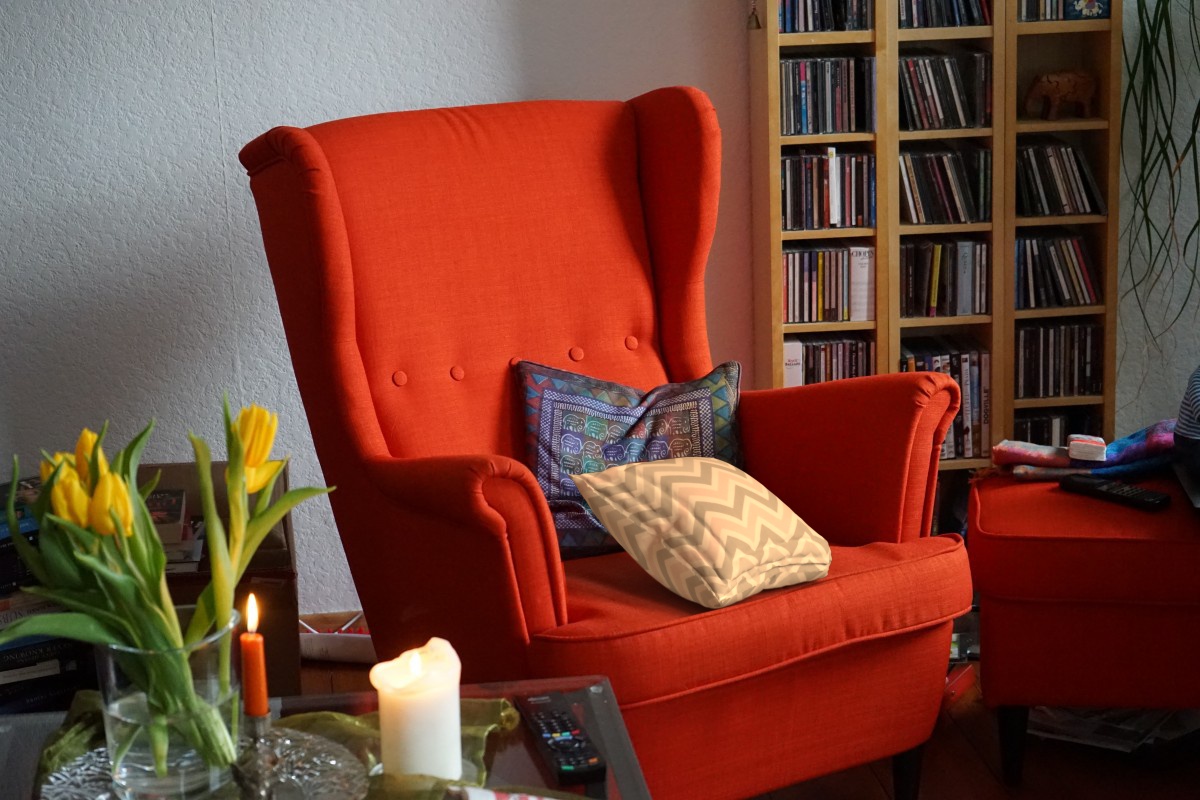} \\
	\includegraphics[width=0.49\linewidth]{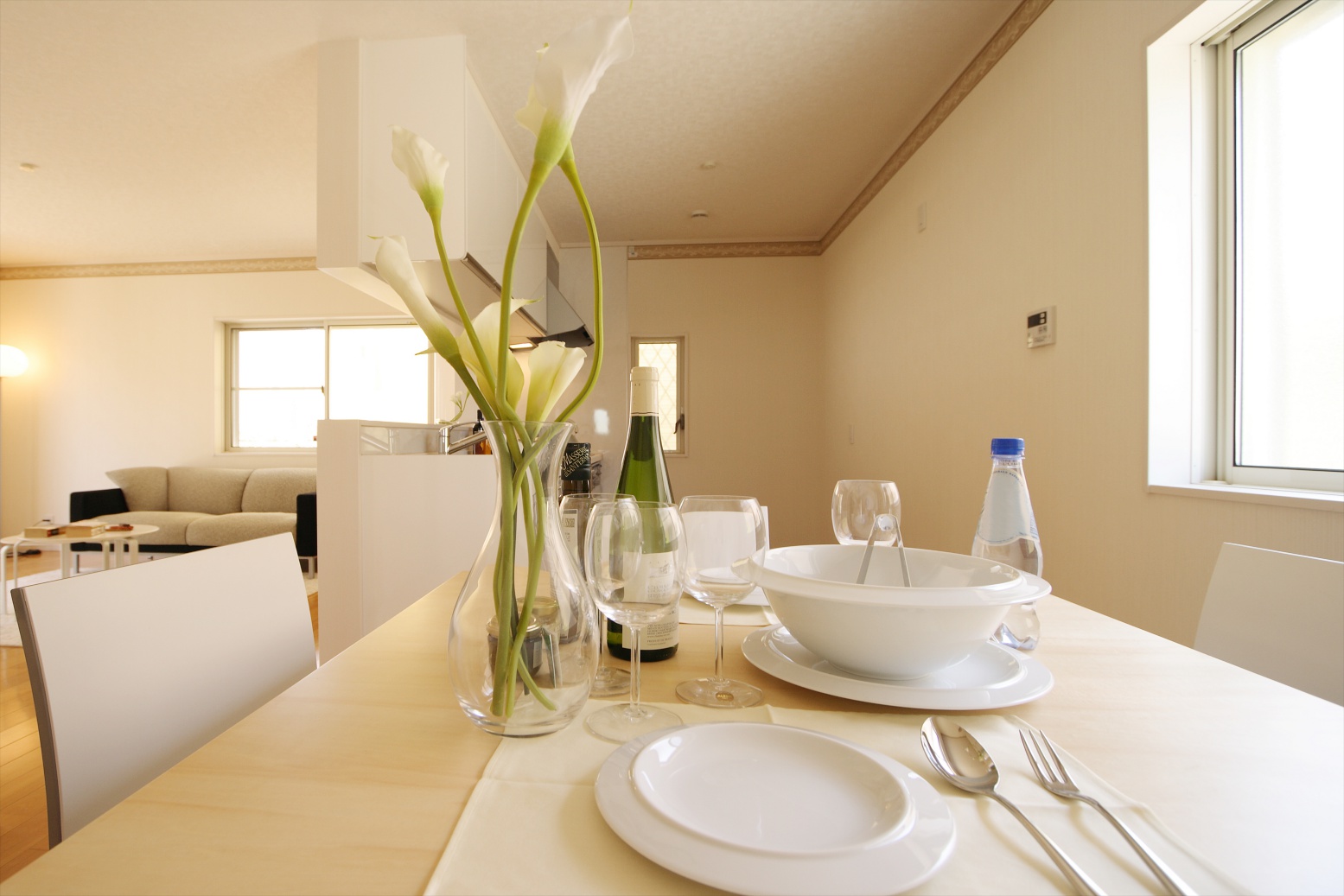} &
	\includegraphics[width=0.49\linewidth]{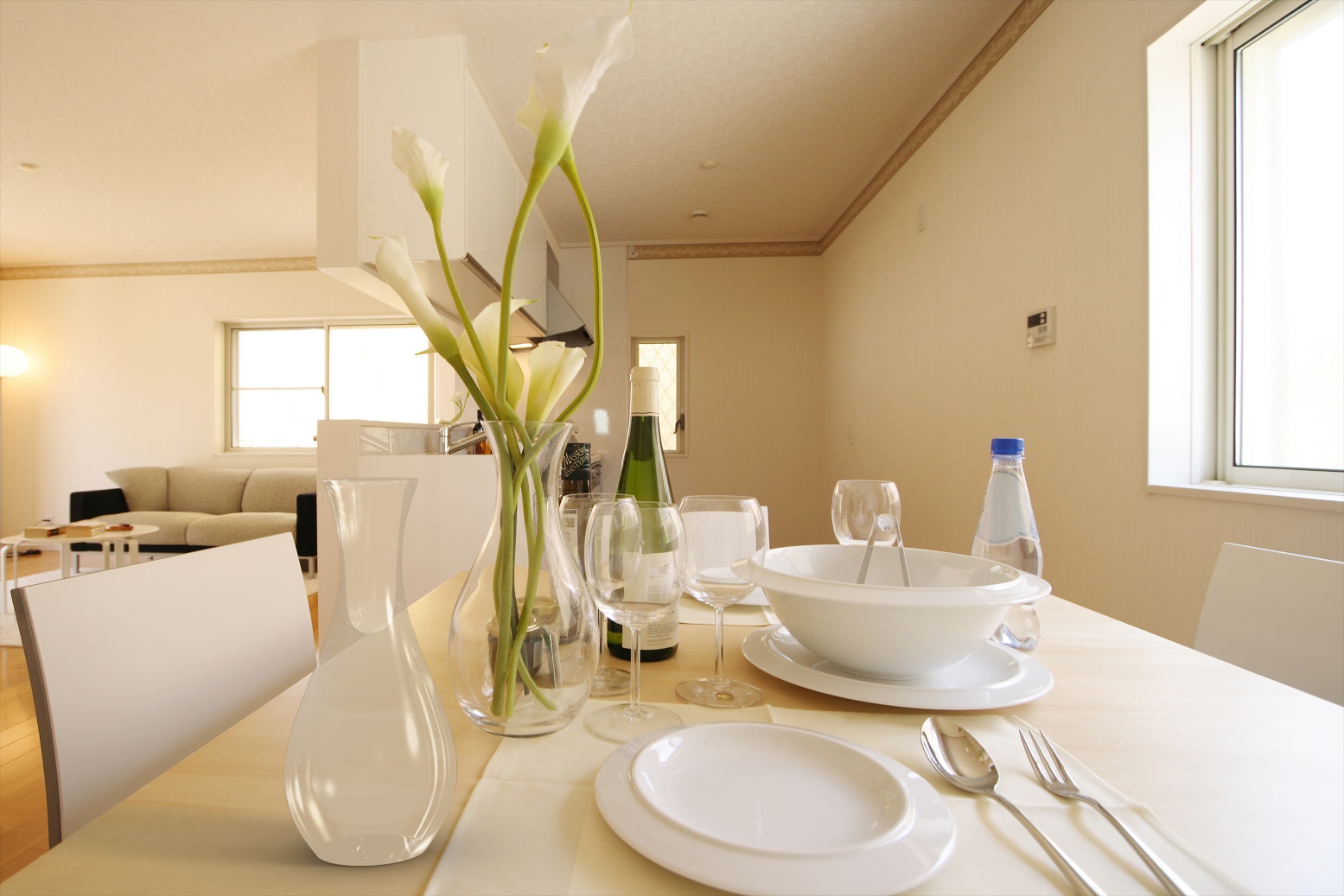} \\
	\end{tabular}
	\caption{Object relighting on a variety of generic stock photos downloaded from the Internet. More results on such out-of-dataset images are provided in supplementary material.}
	\label{f:out-of-dataset}
\end{figure}

\subsection{Qualitative evaluation}

Fig.~\ref{f:qualitativevsgt} shows two representative qualitative comparisons to the non-parametric approach of Gardner et al.~\cite{gardner-sigasia-17}. For each example, the estimated environment map and a render of diffuse and glossy objects are shown. To highlight the difference between parametric (ours) and non-parametric (\cite{gardner-sigasia-17}) lighting representations, the objects are lit by the lighting estimated at the center of the scene. Even though our representation is much more compact than that of \cite{gardner-sigasia-17}, the rendering results are either qualitatively similar (first column) or better match ground truth (last column) than \cite{gardner-sigasia-17}. 

We also compare rendered and real versions of the same object in fig.~\ref{f:bunnies}. For each image, a real bunny model was placed in the scene. The images were acquired using a regular smartphone camera in order to further validate the robustness of the network with out-of-dataset samples. From the network estimation, we relight a virtual model of the same bunny and insert it close to the real one. Although the real and rendered bunnies slightly differ in appearance, the overall comparison demonstrates our method's ability to recover correct light positions and intensity (more results on similar scenes in supplementary material).

Finally, in fig.~\ref{f:out-of-dataset}, we demonstrate our method's ability to generalize to stock photos, including some that have undergone artistic processing, and still produce realistic results.

%% file: discussion.tex
\section{Conclusion}
\label{sec:discussion}

We have presented a method to estimate parametric 3D lighting---comprised of discrete light sources with positions, area, color and intensity---from a single indoor image. Our lighting model allows us to render incident illumination at any location in a scene; this is critical for indoor scenes which often have localized lighting that cannot be accurately modeled by global lighting estimates. We train this method end-to-end using a differentiable parametric loss based on an environment map representation. We have demonstrated that our method is robust and significantly outperforms previous work in terms of lighting accuracy and allows for realistic virtual object insertion where objects are lit differently based on their insertion point.

Our work offers a number of directions for future exploration. First, while our lighting representation is 3D, using it requires a 3D scene reconstruction to specify locations where to estimate lighting and to compute light visibility. Second, our lighting model assumes diffuse area light-like sources and cannot model directional lights or focused light beams. Extending this model to handle more general sources would generalize it to more indoor scenes. Finally, our method focuses solely on estimating scene illumination. However, lighting is only one of many scene properties that affect its appearance. We hypothesize that jointly reasoning about scene appearance---including geometry, materials, and illumination---could improve overall accuracy. 
